\documentclass{article}

\usepackage{arxiv}
\usepackage[utf8]{inputenc}
\usepackage[T1]{fontenc}
\usepackage[authoryear,round]{natbib}
\usepackage{hyperref}
\usepackage{graphicx}
\usepackage{booktabs}
\usepackage{authblk}
\usepackage{xcolor}
\usepackage{array}
\usepackage{caption}
\usepackage{float}

\renewcommand{\headeright}{Preprint $\cdot$ \textit{Phenomenology and the Cognitive Sciences} (2025)}
\renewcommand{\undertitle}{arXiv:2407.10735}
\renewcommand{\shorttitle}{Transforming Agency}

\AtBeginDocument{%
  \fancyhf{}
  \lhead{\scshape\footnotesize Transforming Agency}
  \rhead{\scshape\footnotesize Preprint $\cdot$ \textit{Phenomenology and the Cognitive Sciences} (2025)}
  \cfoot{\thepage}
  
}

\hypersetup{
  colorlinks = true,
  linkcolor  = blue,
  citecolor  = blue,
  urlcolor   = blue,
  pdftitle   = {Transforming Agency: On the mode of existence of large language models},
  pdfauthor  = {Xabier E. Barandiaran and Lola S. Almendros},
  pdfkeywords= {Transformers, Large Language Models, Agency, Autonomy,
                Interlocutor Automata, Human Machine Interaction}
}

\title{Transforming Agency\\[0.4em]
  \large\textit{On the mode of existence of large language models}}

\author[1,*]{Xabier E. Barandiaran}
\author[2]{Lola S. Almendros}

\affil[1]{IAS-Research Centre for Life, Mind, and Society, Dept.\ Philosophy\\
  UPV/EHU, University of the Basque Country, Donostia (Spain)\\
  \url{http://xabier.barandiaran.net} \quad
  \href{mailto:xabier.barandiaran@ehu.eus}{xabier.barandiaran@ehu.eus}\\
  ORCID: \href{https://orcid.org/0000-0002-4763-6845}{0000-0002-4763-6845} \quad
  $^*$ Corresponding author}

\affil[2]{Institute for Science and Technology Studies, University of Salamanca (Spain)\\
  \href{mailto:lola.s.almendros@gmail.com}{lola.s.almendros@gmail.com} \quad
  ORCID: \href{https://orcid.org/0000-0002-1414-0827}{0000-0002-1414-0827}}

\date{}

\begin{document}

\maketitle

%% --- Abstract ---
\begin{abstract}
This paper investigates the ontological characterization of Large Language Models (LLMs) like
ChatGPT. Between inflationary and deflationary accounts, we pay special attention to their
status as agents. This requires explaining in detail the architecture, processing, and training
procedures that enable LLMs to display their capacities, and the extensions used to turn LLMs
into agent-like systems. After a systematic analysis we conclude that a LLM fails to meet
necessary and sufficient conditions for autonomous agency in the light of embodied theories of
mind: the \emph{individuality} condition (it is not the product of its own activity, it is not
even directly affected by it), the \emph{normativity} condition (it does not generate its own
norms or goals), and, partially the \emph{interactional asymmetry} condition (it is not the
origin and sustained source of its interaction with the environment). If not agents, then \ldots\
What are LLMs? We argue that ChatGPT should be characterized as an interlocutor or linguistic
automaton, a \emph{library-that-talks}, devoid of (autonomous) agency, but capable to engage
performatively on non-purposeful yet purpose-structured and purpose-bounded tasks. When
interacting with humans, a ``ghostly'' component of the human-machine interaction makes it
possible to enact genuine conversational experiences with LLMs. Despite their lack of
sensorimotor and biological embodiment, LLMs textual embodiment (the training \emph{corpus}),
digital extended interface embodiments, and resource-hungry computational embodiment,
significantly transform existing forms of machine automatism and human agency. Beyond assisted
and extended agency, the LLM-human coupling can produce \emph{midtended} forms of agency,
closer to the production of intentional agency than to the extended instrumentality of any
previous technologies.
\end{abstract}


%% --- Citation ---
\section*{Citation}

This paper is pending review in a journal. Meanwhile, please reference as:

\begin{itemize}
  \item Barandiaran, X.~E., \& Almendros, L.~S. (2024). \textit{Transforming Agency. On the
    mode of existence of Large Language Models} (arXiv:2407.10735). arXiv.
    \url{http://arxiv.org/abs/2407.10735} Published in \textit{Phenomenology and the Cognitive
    Sciences} \url{https://link.springer.com/article/10.1007/s11097-025-10094-3}(2025).
\end{itemize}

%% --- Acknowledgments ---
\section*{Acknowledgments}

XEB and LSA acknowledge IAS-Research group funding IT1668-22 from Basque Government, grants
PID2019-104576GB-I00 for project Outonomy, and PID2023-147251NB-I00 for project Outagencies
funded by MCIU/AEI/10.13039/501100011033. LSA also acknowledges the University of Salamanca's
Margarita Salas Postdoctoral Grant Program within Spain's Recovery, Transformation, and
Resilience Plan and Next Generation EU\@. Special thanks to Steve Torrance, Miguel Aguilera,
Enara Garc\'ia, Marta P\'erez-Verdugo, Ekai Txapartegi, Ana Valdivia, Matthew Egbert, and
Antonio Calleja-Lopez for careful and insightful revision and comments to early manuscripts of
this paper.

\tableofcontents

\newpage

%% ================================================================
\section{Introduction}

The recent emergence of Large Language Models (LLMs hereafter) \citep[see][]{brown2020} with
their wide availability\footnote{Online services like Gemini, Claude, or, more prominently,
ChatGPT, and the free/open source alternatives that can also run locally, like DeepSeek, Qwen,
LLaMa, Mixtral, etc.\ are making these technologies massively available, not only as direct
conversational bots but also, and importantly, as integrated assistants or agency boosting
systems into different applications.} and their human-like generative capabilities are
(re)opening the question around the ontological status of Artificial Intelligence. Are these
systems genuinely intelligent? Do they possess mindful capacities? The responses are often
polarized \citep{mitchell2023}. Inflationary views (fuelled by research enthusiasm and
commercial interest alike) tend to amplify AI properties, assimilating or approximating them to
the human (and the superhuman). Deflationary views (typically trying to mitigate the harms of
inflationary marketing), tend to downplay capacity attributions, and bring AI systems closer to
dumb mathematical or mechanical devices. \emph{Deflationary} categorizations typically revolve
around treating LLMs as statistical processors, ``stochastic parrots'' \citep{bender2021}, a
``blurry JPEG of the web'' \citep{chiang2023}, ``lumbering statistical engine for pattern
matching'' \citep{chomsky2023}, ``illusory thinking'' \citep{shojaee2025} or simply
``bullshit'' \citep{hicks2024}\footnote{Although all the cited deflationary accounts of what
LLMs do provide critical arguments to defend their position, the term ``bullshit,'' as used by
Hicks et al., is more technical and better defended than it might appear at first sight. Hicks
et al.\ adopt Frankfurt's \citeyearpar{frankfurt2005} explanation of ``bullshit talk'' as
discourse that disregards truth without directly aiming to mislead the receiver or intentionally
conceal it, as lying does. They show how this analysis applies to LLM outputs as well.}. The
most \emph{inflationary} characterizations range from considering LLMs as ``human-brain
equivalents'' \citep{ge2023}, ``genuine authors'' and ``accountable'' entities
\citep{miller2023}, up to a ``fully sentient person'' \citep{lemoine2022}. Somewhere in the
middle stand more technical characterizations like ``artificial reasoners'' \citep{wei2023},
``learners'' \citep{brown2020}, ``general pattern machines'' \citep{mirchandani2023}, ``sparky
artificial general intelligence'' \citep{bubeck2023} or simply ``language models'' with the
slippery temptation to be turned into ``world models'' \citep{liK2023}.

An increasing danger of some deflationary views of LLMs is that, from their point of view, most
of the risks can be attributed to the influence of misguided inflationary conceptions. It is
often assumed that these can be mitigated if inflationary views are conclusively shown to be
wrong: ``Behind the smog of the hype and the marketing'' the argument goes ``there is no genuine
intelligence or understanding behind LLMs, they are simple statistical processors, the only
problem (besides their energy consumption and biases) is that other humans take them at face
value''. Moreover, the argument continues, ``if we treat them as the stupid machines they are'',
the conclusion follows, ``even the issues of bias and energy should fade away''. Or, as
\citet{chomsky2023} conclude, ``Given the amorality, faux science and linguistic incompetence of
these systems, we can only laugh or cry at their popularity'' (2023). The real capacity of LLMs
is thus left disregarded by such deflationary views, both as a potential risk to society and as
a genuine source of positive sociotechnical transformation that needs to be more deeply thought
out. If we do not fine-tune our conceptualization of what LLMs are, we will not be able to
properly analyze, evaluate, stir, and communicate their impact.

There are many ways of assessing this ``real'' power and its impact. Some are historic and
socio-economic \citep{pasquinelli2023}, or socio-ecological \citep{crawford2021}, or even
existential \citep{bostrom2017,christian2021,russell2019}. But little attention has been put on
critically analyzing LLMs from the point of view of ``agency''. Despite the widespread academic
consensus on the lack of conscious or sentient capabilities of LLMs, their status as agents is
often uncritically assumed or proclaimed both in philosophy \citep{floridi2023} and the industry
\citep{openai2025b}. And there are two good reasons to strengthen accuracy on agentive
attributions to LLMs: a) mindful capacities did not arise in nature as a result of chess playing
but of the evolution of agency \citep{barandiaran2008,sterelny2001,tomasello2022} and thus
granting LLMs the status of agents means assuming them on a path towards such properties, and b)
achieving autonomous agency is one of the next big things in AI \citep{wang2023} with widespread
support from Google, Apple, and Microsoft (including OpenAI) recently marketing their AI
products as ``agents'' \citep{holmes2024,odonnell2024}; a point that became particularly
relevant since the recent launch of ChatGPT Agent \citep{openai2025b}.

In the next section, we contextualize the problem of the ontological status of LLM in terms of
its current capabilities and limitations as expressed on different benchmarks. We show how
transformer technologies are breaking down the solid distinctions between the human and the
engineered. The \emph{mode of existence} of these technologies is, however, not limited to
their performance. In order to clarify it, we follow a threefold approach. First, in section~3,
we ground our analysis in the technical reality of these systems by focusing on the internal
workings of ChatGPT-like LLMs: their architecture, processing flow, training procedures, and
the extensions used to create agent-like behaviours. A clear technical understanding is the
necessary foundation for any meaningful ontological assessment. Second, with these technical
details at hand, in section~4, we systematically evaluate their status as agents from the
perspective of contemporary embodied and enactive theories of mind. We conclude that they fail
to meet the core conditions for autonomous agency. Third, in section~5, we argue that this
negative conclusion is insufficient; it compels us to address the crucial question: If not
agents, then what are they? Based on the technical specifications of section~3, and building on
embodied and enactive approaches to cognition, we characterize them as \emph{interlocutor
automata}, capable of bringing digital textual bodies to conversational life in interaction with
humans, with the capacity to deeply transform human agency. We finally conclude and discuss the
implications of our approach.

%% ================================================================
\section{When computers can}

Benchmarks, particularly when out of reach for the available technology, have often helped to
reach agreement on the (lack of) capacities of AI systems. Now, many defend that the Turing Test
is outdated \citep{bayne2023,biever2023,srivastava2023,tikhonov2023}. Generic conversations with
LLMs are indistinguishable from those we can enjoy with other humans \citep{jones2024}. More
systematic variations of the Turing Test, directed at capturing the capacity of AI to display
common sense, like the Winograd schema challenge \citep{levesque2012}, have been declared
obsolete \citep{kocijan2023}. More sophisticated common sense reasoning tests like Winogrande
\citep{sakaguchi2019} and HellaSwag \citep{zellers2019}, specifically designed to be
particularly hard for LLMs, have also been passed \citep{gemini2023,jiangAQ2023,openai2023}.
Moreover, by 2023, LLMs, like GPT-4 already exhibited, according to their creators,
``human-level performance'' in a wide variety of professional and academic exams
\citep[p.~6]{openai2023}, or, like Google's Gemini, improving GPT-4 in many benchmarks
\citep{gemini2023}, could outperform humans on multitask language understanding tests
\citep{hendrycks2021}.

Current proprietary frontier systems---GPT-4o \citep{openai2024a}, the GPT-o3 and o4-mini
\citep{openai2025a}, Claude 4 family \citep{anthropic2025}, and Gemini 2.5 Pro
\citep{googledeepMind2025}---now reach or surpass human-level accuracy on many reasoning and
commonsense suites. Open-source counterparts have rapidly closed the gap: KIMI K2
\citep{moonshot2025}, Llama 4 \citep{metaai2025}, Mixtral of Experts and its successor
Magistral \citep{jiangAQ2024,mistral2025}, DeepSeek-R1 \citep{deepseek2025}, or Qwen 2.5
\citep{qwen2025} all report benchmark scores within a few percentage points of their
closed-weight peers, while offering transparent and reproducible alternatives for academic
research.

In concrete knowledge domains (like medicine) frontier models since 2023 often outperform
average specialists at specific tasks \citep{guo2023,singhal2025,vanveen2024} and, perhaps not
so surprisingly, can imitate philosophers with hardly distinguishable snippets
\citep{schwitzgebel2024}. More recently, methodological innovations on LLM's self-guided
``reasoning'' \citep{lightman2023,madaan2023a,wei2023,yao2023}, and the resulting increase of
inference-time compute, have pushed state-of-the-art scores by even higher margins
\citep{deepseek2025,openai2024b}. A significant boost in benchmarking has also resulted from
agentic extensions and tool use of LLMs \citep{openai2025b}.

This rapid progress has accelerated the obsolescence of existing benchmarks, forcing the latest
1M dollar prized Artificial General Intelligence benchmark (ARC-AGI) to be updated with a more
challenging successor ARC-AGI2 in March 2025 \citep{chollet2025}, and an even more challenging
and agent focused ARC-AGI3 in July 2025\footnote{\url{https://three.arcprize.org}}.

However, aggregate leaderboard numbers can hide ``pockets of unpredictable failure'' and thus
overstate real-world reliability \citep{burnell2023}. There are arguable limitations of current
models, particularly in relation to some abstract reasoning capacities like compositionality
\citep{dziri2023}, multistep reasoning \citep{sprague2023}, or complex planning
\citep{valmeekam2023a}; whose mastery, by the way, is often rare among
humans\footnote{Most recently, Apple's \emph{The Illusion of Thinking} study \citep{shojaee2025}
attracted extraordinary media attention in an increasingly polarized context. Subsequent
analyses, however, have contested the procedure and conclusions, attributing much of the
reported failure to evaluation artefacts and token-budget limits \citep{lawsen2025}.}.
Large-scale studies show that scaling and instruction fine-tuning might not \emph{fully}
eliminate these weaknesses and could even make errors harder to predict \citep{zhouL2024}.
Moreover, it is still possible, although increasingly harder, to find tasks in which humans score
high, without specific expertise, yet remain hard for LLMs, as shown by the ARC-AGI-2 and 3
benchmarks \citep{chollet2025}. And yet, what is certain is that we are facing the development
of complex technologies that perform operations whose results are very similar to those requiring
high levels of intelligence in humans. This circumstance translates into a growing
undifferentiation between the human and the engineered. Conceptual divides that were once sharp
and fixed are starting to melt and move. These advances force us to re-organize conceptual and
ontological commitments regarding minds, machines, and agency.

Ever since the very conception of modern computers as universal Turing machines, the possibility
of instantiating human, or super-human, level intelligence was at stake \citep{turing1950}. At
some point, this race between mind and machines settled down. Machines could (out)perform humans
provided that the domain of interaction was rule-based, constrained, and limited, so that humans
could program machines specifying the computational procedures required to carry out the task.
Machines, however, were left with genuine mindfulness out of reach. Real-world tasks such as
open conversations, a trip to the grocery, creative writing or subtle comforting humor were only
within reach for us. The (human) mind could not be reduced to rule-following or explicit
reasoning capacities but emerged, instead, out of sub-rational skillful embodied interactions
that could not be implemented into machines \citep{dreyfus1992}. Elephants, after all, don't
play chess, but their mental life is rich and complex, nothing like computers would ever be able
to accomplish \citep{brooks1990}. On the other hand, computers could play chess but not pass the
Turing Test, they could imitate and outperform humans in specific rule-based scenarios but not on
the open field of language games and skilled conversations fuelled by a \emph{common-sense}: an
embodied sub-symbolic mesh that was claimed to resist rational, explicit, operationalization
\citep{johnson2002,varela1991}.

Transformers have come to break this cease-fire between minds and machines\footnote{Although
some pre-transformer successes were already anticipatory of the progress that AI development was
about to suffer. First it was GO, an open-ended, combinatorially explosive game that cannot be
played by but by intuition in a manner that GO fans consider that is a pure expression of the
player's soul. And second, perhaps more importantly, in playing different computer games, using
human controls (e.g.\ first-person visuomotor feedback), and without knowing or encoding the
game rules in advance \citep{schrittwieser2020}.}. LLMs can carry out context-sensitive
translations, they can explain humor and jokes including interpreting what humans take to be
funny images \citep{hessel2023}, albeit with notable limitations, can learn from few or a single
example or instruction \citep{brown2020}, or engage on reasoning chains ``creatively''
\citep{wei2023}. Moreover, LLM technology and transformer architectures are being applied to
multiple sensorimotor modalities both in real-world physical robots through VLA (Vision Language
Action) models \citep{brohan2023,collaboration2023}, Q-learning enhanced LLMs
\citep{chebotar2023}, or directly applied to visual and sensorimotor tasks
\citep{bousmalis2023,reedS2022}. These expansions of LLMs might succeed out of the
text-image bound domain into physically enacting sensorimotor correlations and learning skillful
coping with the physical world \citep{xiangJ2023}. An approach that brings new generation AI
systems much closer to traditionally ``AI-skeptic'' embodied and situated approaches to mind and
cognition \citep{chemero2009,dreyfus1992,johnson2002,varela1991}. It is, however, too early to
judge the success of embodied robotic implementations of new generative technologies. On the
contrary, there is a suite of text-interfaced LLMs (ChatGPT, LLaMa, Gemini, Claude, Mistral,
DeepSeek, Qwen, etc.), providing first-hand experience for millions of people, defining the way
we are relating to LLM technologies and transforming the digital (and non-digital) human
environment.

The best way to avoid alienation is not to feed inflationary positions or to join the
deflationary ranks, but to find the right ontological categorization for these systems; or, to
say it with \citet{simondon2017}, to identify the \emph{mode of existence} of technical
systems. Like the case of consciousness or sentience, we cannot leave the answer to ``social
relationism''; i.e.\ to a mere social contingent convention on what type of systems deserves
which treatment (for a detailed argument against social relationism and AI see \citealt{torrance2014}).
A proper understanding of what LLMs \emph{are}, requires delving deep into their concrete
structure, operations and coupling with their \emph{milieu} (humans and other machines). Can,
and should, we take them for agents? Are they intelligent? If not, what are they? What is the
best way to conceptualize them? The answer has important implications in the field of ethics and
legal studies \citep{bertolini2022,clowes2024,coeckelbergh2021,fourneret2020,mabaso2021}, but
also on the social adoption of these technologies, our collective awareness of their limits and
potentialities. We need conceptual resources to organize our experience and interactions with
ChatGPTs and their place in our sociotechnical world.

Understanding in some detail how GPT and other LLMs are trained and how they function is of
fundamental importance to characterize their ``nature'', genuine capabilities and possible
implications. In the next section, we provide an explanation of how GPT works, with the goal of
contrasting its actual functioning with some ontological attributions, particularly its agentive
capacities. A detailed understanding of GPTs functioning will also help characterize its mode of
existence and the ways in which it can potentially transform human agency\footnote{The next
section provides a detailed technical overview of how LLMs function, intended to ground the
subsequent philosophical analysis in Sections~4 and~5. Readers already deeply familiar with the
architecture and training procedures of transformer models may choose to skim this section or
proceed directly to the philosophical arguments beginning in Section~4. Note, however, that
specific details will be drawn upon in the later sections.}.

%% ================================================================
\section{How do Large Language Models work?}

Large Language Models (LLMs) are so-called ``artificial intelligence'' systems
\citep{norvig2021}, part of current NLP (Natural Language Processing) technologies, that belong
to the family of ``machine learning'' and the sub-category of ``deep learning'' systems. They
are designed to process and generate ``natural'' language through a large number (on the order
of billions) of processing steps. Transformers \citep{vaswani2017}, in turn, are one kind of
recently very successful type of LLMs, and GPT (Generative Pre-trained Transformer) is a
specific type of implementation of Transformer technology \citep{brown2020,radford2019}. Most
popular LLMs are delivered as a service by a provider. So, for instance, \textit{ChatGPT}, is a
specifically tuned and interfaced version of GPT (and, increasingly, a platform to connect GPT
to other tools and to deliver personalized services with GPT technology)\footnote{On what follows
we shall use the terms ``Transformer'', ``LLM'' and ``GPT'' and ``ChatGPT'' almost
interchangeably, unless specific reference is provided to the concrete model (e.g.\ GPT-2,
DeepSeek-r1) or to aspects of their interface.}.

ChatGPT uses different versions of GPT models to produce human-like text (GPT-4.5, GPT-o3,
etc.). It is a computational language processing system designed to generate sequences of words,
codes or other data (more recently, images) from an input sequence called ``prompt''. Thus,
given a prompt, GPT produces the text that would have been more statistically expected on the
training data. For example, if the sequence ``Elephants don't play'' is entered as a prompt, the
ChatGPT offers ``Elephants don't play \textbf{chess}'' as a response. The system has a heat
parameter that increases less likely variations on the output. So, for instance, the system might
respond to the original prompt with ``Elephants don't play \textbf{video games}'' or could
simply output ``Elephants don't play.''. This basic functioning is what made so popular the
characterization of ChatGPT as simply a complicated auto-complete tool \citep{floridi2023}.

However, the simplicity of the general task of optimizing to predict the following word, and its
recursive iteration, is the key for the emergence of complex capacities in LLMs. Moreover,
optimization alone provides no ground to understand the working of a system, its capabilities
and limits, its mode of existence. Appealing to partial aspects of how they operate, Transformers
are often qualified as stochastic, probabilistic and statistical
\citep{bender2021,chomsky2023,floridi2023}. \emph{Stochasticity} refers to the randomness of how
the final output is ``selected''. \emph{Probabilistic} is used to indicate that this final
decision is taken randomly but on the basis of an assigned probability that is, in turn,
allegedly extracted from the \emph{statistical} properties of the training data. Understood on
its most generous terms, such descriptions are relatively correct but partial and incomplete. It
is possible to imagine a strictly statistical AI that simply computes or extracts conditional
probabilities of all possible output tokens given an input stream. But this simply does not work.
In fact, there could not be sufficient training data on the universe to make such a machine
effective on the basis of pure probabilities or statistics \citep{wolfram2023}. And it is not
what GPT does. As their name indicates, LLMs create \emph{models} of language. That is, they
don't simply store statistical relations or conditional probabilities but instead constitute
compressed and structured engines that process and transform text input in non-linear, highly
interrelated and complex forms.

Before we start with the description of GPTs architecture and processing, it is important to
stress that, in general, the processing blocks and procedures are, technically speaking, complex
\citep{bechtel2010}. They do not make ``sense'' from the point of view of the functional
decomposition of the treatment of the input. Certainly not one that a human might understand or
guess as a reasonable strategy. Admittedly, there is no theory for how and why this architecture
works \citep{wolfram2023}. And yet it works, and we have access to the processing architecture
of GPT-2 \citep{radford2019}, and some details of GPT-3 \citep{brown2020}, to better delimit,
without a possibly full understanding, how the system operates.

In what follows, we will provide a detailed explanation of GPT-3 to the best available
knowledge, as the prototypical LLM and as an example of generative AI more generally. We have
chosen GPT-3 as our foundational case study for several key reasons. First, its release marked a
crucial inflection point\footnote{This is so to the extent that the term ``GPT-3 moment'' has now
been popularized, referring to the breakthrough at which a technological advance converts prior
hypotheses or intuitions into a clearly recognisable technical success.}. It represented a
qualitative leap in generative capability and widespread public adoption. Second, it did set up
the industry standard for other models to follow, becoming the paradigmatic example of generative
AI\@. Subsequent models have increased in size, explored more optimization with novel training
techniques or, as we shall see, multiple parallel specialized transformers and multimodality. But
they largely build upon the core principles of the Generative Pre-trained Transformer
architecture that GPT-3 popularized. A detailed understanding of this particular model provides
the essential conceptual scaffolding for grasping the entire family of these systems, including
their more recent and complex versions, and those developed by other organizations.

%% ----------------------------------------------------------------
\subsection{Architecture and processing}

Figure~\ref{fig:gpt3} illustrates step-by-step the processing of the input text as it goes along
the GPT architecture. We explain each step in detail below.

\begin{figure}[H]
  \centering
  \includegraphics[width=0.55\textwidth]{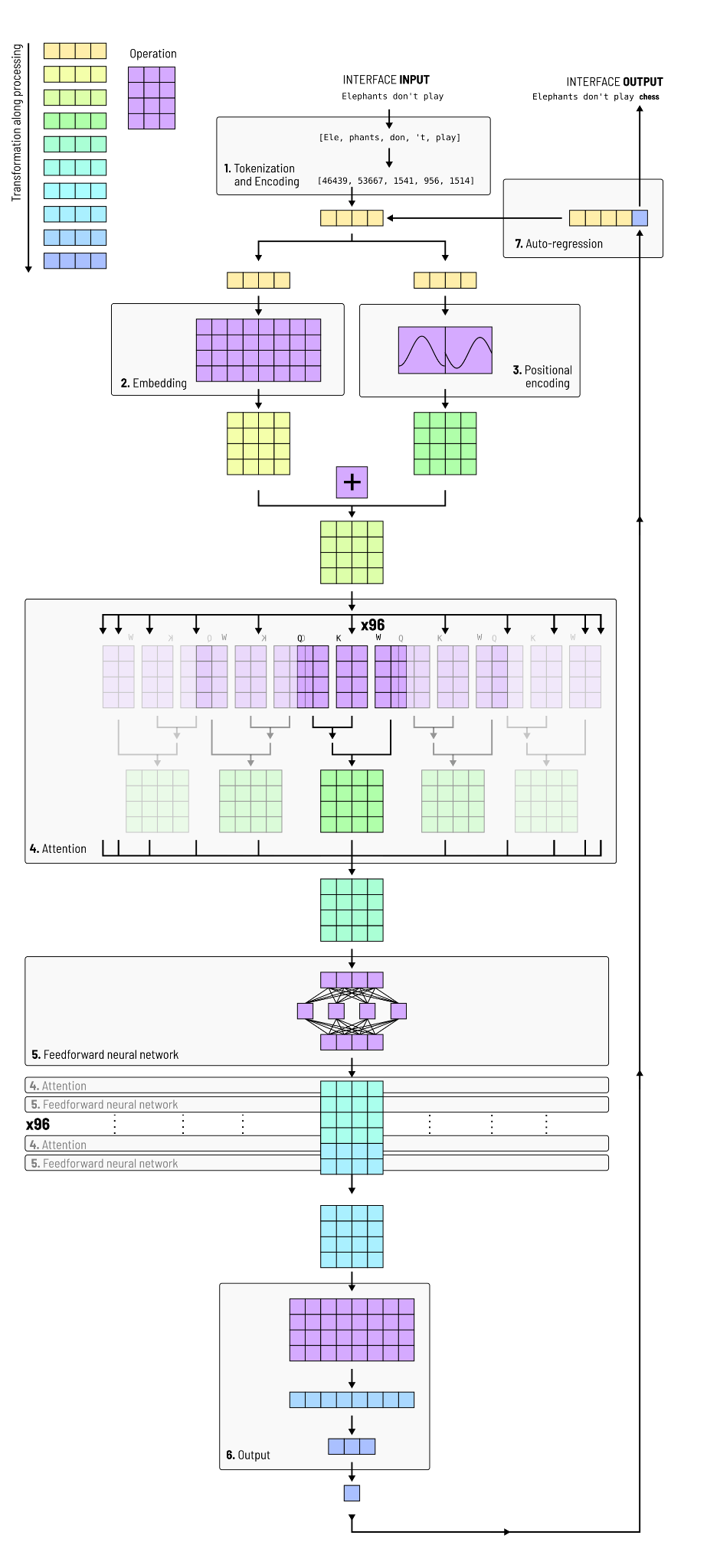}
\end{figure}

\newpage

\begin{figure}[H]
  \centering
  \caption{\textbf{Fig.\ 1.} \textit{Schematic of the GPT-3 processing architecture as a
    standardized reference for LLM functioning.} Computation flows from the top of the diagram
    to the bottom: the input string (shown in yellow) is progressively transformed into
    intermediate matrices whose colour shifts from yellow to green to blue, culminating in the
    selection of a single blue output token. Parameter matrices (marked in purple) act on these
    intermediate matrices (they remain fixed during processing but are configured and updated
    during pre-training and training, see sect.~3.2). (1)~The input string is tokenised and
    numerically encoded. (2)~Tokens are embedded as 12288-dimensional vectors or ``semantic
    space'' defined for a 50257 token (word) vocabulary. (3)~A positional encoding is added to
    each vector to preserve word order. The resulting matrix passes iteratively through 96
    sequential transformer blocks, each involving the processing through (4)~96 parallel
    self-attention heads and (5)~a high dimensional feed-forward neural network. (6)~A
    projection back into vocabulary space followed by softmax normalisation yields a probability
    distribution over the 50257 possible tokens, from which the next token is sampled.
    (7)~The sampled token is added to the original input (returning to step~1), and the procedure
    repeats auto-regressively until an end-of-sequence token is produced or the maximum length is
    reached. The text body provides a detailed explanation of each processing step.}
  \label{fig:gpt3}
\end{figure}

\textbf{1.\ Tokenization and encoding}: The first operation that takes place as we enter text
into a LLM like ChatGPT is \textit{tokenization}. The input stream is chopped into \textit{tokens}
(small syllable-like or small word text chunks, including punctuation marks). On average, each
token is about 3/4 of a word in English, with a mean of 4 characters per
token\footnote{It is possible to work with Tokenizer to understand better this procedure:
\url{https://platform.openai.com/tokenizer}}. Nevertheless, we are going to use the terms
``word'' and ``token'' interchangeably. Then each token is encoded numerically. So for example,
``Elephants don't play'' is tokenized into five tokens: [Ele, phants, don, 't, play]. And then
each of them is assigned a predefined number out of the complete vocabulary of 50,257 tokens,
and the sequence is converted into an array of numbers: [46439, 53667, 1541, 956, 1514].

\textbf{2.\ Embedding}: Once tokens are represented numerically, these numbers are mapped into a
high dimensional relational space. This process is called embedding. Embedding already implies a
huge transformation of the input with previous ``knowledge'' of how tokens (or words) relate to
each other. Some of these relationships can be considered ``semantic'' or ``syntactic'' by
capturing higher-order relational properties between words. Some dimensions or combinations of
dimensions might be thought of as abstract conceptual properties (e.g.\ being a grammatical
subject or being an animal). What defines the conceptual content of each dimension is not an
arbitrary label into it, but purely relational ``spatial'' properties. For instance, animal names
will appear close to each other, also grammatical subjects, etc. The embedding space of GPT-3 is
of 12,288 dimensions \citep{brown2020} and a position within that space is pre-encoded for
50,257 tokens\footnote{We provide token and matrix size details here with two specific goals:
a) to illustrate the scale of these systems---a point that will turn relevant in section~3.3---and,
b) to track the matrices that are combined at different steps of the process; the size of the
matrix is thus provided as a label or identifier. The specific size is, of course, not relevant,
but helps track the matrix transformations along the text and in relation to Figure~1.}
\citep{radford2019}. One way to understand this (with the risk of anthropomorphizing) is to say
that GPT3 has the capacity to situate 50,257 words\footnote{We will use the terms token and
words interchangeably for a more intuitive grasp of the functioning of the system. It is
difficult to be strictly rigorous here because the concept of word itself is ill-defined, with a
regular human knowing approximately 10k word families \citep{brysbaert2016}.} in a 12,288
dimensional ``conceptual'' space (or along 12,288 ``properties''). So, for instance, the tokens
``cat'' and ``tiger'' will be close to each other in many dimensions but will be relatively
spaced in ``size'' and ``habitat''. This means nothing other than the token ``cat'' being closer
to ``laptop'', ``dog'' and ``watermelon'' on a given dimension (that we could interpret as
``size'') and closer to ``Roomba'', ``television'', ``sofa'' and ``living-room'' in another
(that we might interpret as ``habitat'').

Embeddings already embody an important part of the ``knowledge'' of an LLM, and its production
is part of the overall training procedure of GPT\@. The result of applying the embedding
function to the input stream is a matrix of 2048$\times$12288 that is itself called
embedding\footnote{It is a regular practice to use the term embedding to name the matrix that
will be processed along the whole transformer. But this might lead to confusion. Although the
matrix maintains the same size, and the end result will be transposed back into tokens (see
latter), the successive operations carried out over this matrix distort its original
interpretation so much that we find it confusing to keep calling it ``embedding''. We should use
the term ``matrix'' instead and identify it by its size.}. The number 2048 (for GPT3, the size
in current models is much higher) indicates the maximum size of the input stream in tokens. It
sets the upper limit of how much ``context'' (previous conversation, additional information or
maximum input provided) can the system handle. Although the input might just be of 5 tokens and
the resulting matrix 5$\times$12288, from now on, for convenience we will assume that the matrix
being processed is this 2048$\times$12288 maximal sized one.

\textbf{3.\ Positional encoding}: Not only the set of words composing the input matters, their
order also does. It is not the same to say that ``John Searle invented the Chinese Room'' than
saying that ``The Chinese Room invented John Searle''. So each word/token embedding will also be
transformed to incorporate positional information. The positional encoding in the form of a
unique sine and cosine function output is added to the word embedding. A wave signature is added
to each embedding array that is unique to a specific position and can be exploited by later
processing to identify the position of that token on the original input stream. This produces a
new matrix of 2048$\times$12288 with all the 2048 input tokens in one dimension and their
embedding + position on the other 12,288. The combination of embedding and positional encoding
will now be processed through a sequence of processing blocks, like a factory line. Each block
consists of a set of operations that include primarily: attention, addition and normalization, and
feed-forward neural network processing. GPT-3 transforms the input matrix through 96 such
blocks.

\textbf{4.\ Attention mechanism}: This is the most innovative of all the steps on the LLM
revolution and characterizes transformers as a specific type of LLMs \citep{vaswani2017}.
Attention layers have permitted an increase in LLM size and efficiency due to their capacity to
parallelize processing during learning and execution time. They permit to explore a wide range of
correlation dependencies over the input data, in a highly scalable manner; improving upon other
architectures aimed at processing relationships between text elements in an input (e.g.\ retaining
a working memory of the past words in a paragraph), like Recurrent Neural Networks or Long-Short
Term Memory Networks.

Attention mechanisms basically compute how important are some tokens in a stream (and how other
tokens can be ignored) but also how important are the relationships between tokens in the input
stream. This might pick out short and long distance relationships, chopping the input stream into
different chunks. Some of these attentional relationships might capture grammatical connections,
like the verb whose subject is far behind it in a sentence. Others might capture instructional
(e.g.\ the relationships between different steps of a recipe) or narrative structures (like the
unfolding of plot and the connection between characters through time). This is transiently
expressed as a matrix in which all the tokens are valued in relation to all other tokens relating
``everywhere, all at once'', in what would horrify Bergson as a geometrization of duration. By
computing in parallel 96 attention mappings of this kind, transformers avoid the computational
bottleneck of recurrent sequential processing. In short, attention mechanisms make it possible to
be sensitive to different contextual scales. GPT-3 processes 96 attention
heads\footnote{Not to be confused with the 96 blocks. Attention heads run in parallel inside each
block. Block processing takes place in a sequence, the matrix that results from the
transformations of block-$n$ are the input to block-$(n+1)$. Attention heads process 96 copies of
the input matrix in parallel, and then all 96 are added and normalized into a single matrix that
is then further processed. See Figure~1.}, that means that it processes (and later combines) 96
different ways of relating the input sequence to each other. What exactly each of these heads
``really pay attention to'' is partially unknown.

\textbf{5.\ FeedForward network}: The next step involves passing the matrix through a FeedForward
Neural Network (FFNN) and it involves an important expanse on the dimensionality of the
processing and the non-linear interaction between all the components of the matrix; relating
``everything, all at once''\footnote{This is the most unknown or unexplainable part of the
transformations that the input suffers. The operations are simple and vaguely inspired on how
natural neuronal networks function. But what exactly are the structural changes that take place
and what they correspond to in terms of humanly explainable linguistic or cognitive operations
are fundamentally unknown. And might inevitably remain so.}. The feedforward network consists of
3 layers. The input layer is the matrix itself, the hidden layer expands its dimensionality and
the output layer reduces it back to the original size. All the nodes of the first layer are
connected to all the second, and all the second to the third (but not between themselves nor
backward, thus the name FeedForward). The connections are weighted, so that different
relationships between value projections can have different weights and amplify or reduce the value
of each signal into the next layer. This is then processed by the nodes of the next layer through
a nonlinear function. In principle, FFNNs can compute any function \citep{siegelmann1995}. In
this case, they could be thought of as a computer inside a computer that could simulate different
types of programs\footnote{Although they are not technically ``Turing complete'' due to deficits
in recursion and memory.}, with the benefit of being programmable in an unsupervised manner (see
training section below). The weights and the parameters of the nonlinear function have
traditionally been understood as the ``place'' where ``knowledge'' is encoded
\citep{churchland1990,rumelhart1987}. So, for instance, the fact that GPT responds with ``chess''
to the sentence ``Elephants don't play'', instead of simply ``.'' or ``basketball'' is not
something to be found on the original embedding, where certainly ``chess'' is an option close to
``play'' but certainly not the closest. Traces of Brook's famous paper ``Elephants don't play
chess'', its poetic ``value'', and other contextual elements (e.g.\ talking about AI and the
notable role that chess played in its history) can explain the final output.

Attention takes over 30\% of parameters of the model and FF about 70\% in the largest GPT-3 175
Billion parameter model \citep{huben2023}. After embeddings, positions and attentional processing
has taken place, FFNN processing ``elaborates'' relations between tokens applying to them the
knowledge that was acquired through the training process. But arrays of the resulting matrix can
hardly be understood as directly relating to the original tokens anymore. Less so in subsequent
transformations, since the matrix will now be the input to another block that starts again with
its 96 attention heads (different to those of the previous block) and its FFNN processing that
has also specific parameters (weights and biases) in each block. At the end of the process, the
original matrix is severely transformed on its values and is ready to be finally transformed into
the output.

\textbf{6.\ Output}: the 2048$\times$12288 matrix that resulted from the processing of previous
blocks needs now to be converted into a single next token for the original input array. Recall
that the original embedding projected a vocabulary of 50257 words (tokens) into a 12288
dimensional space. A 12288$\times$50257 projection matrix (which is a transpose of the embedding)
now transforms the processed matrix into a score for each token of the vocabulary. The top-$k$
highest scored tokens are separated, and a \emph{softmax} algorithm simply transforms each token
punctuation into a normalized probability that is proportional to its score. Then a final token
is selected according to these probabilities. Visually, this can be likened to a roulette wheel,
where each segment's size is proportional to the token's assigned probability. The selection
process mimics the spinning of this wheel, with the chosen segment indicating the next token in
the sequence.

\textbf{7.\ Auto-regression}: The above sequence of operations is repeated again and again,
adding a new token to the end of the string (e.g.\ a new word to the sentence) until the maximum
number of allowed tokens is reached or, most commonly, an end-of-sequence token is produced by
GPT (a kind of ``halt'' token that is interpreted as a stop). It is possible to continue the
process by reintroducing the input again and adding some more text (like when we add a response
to the conversation). Although the fact that GPT'S ``intelligence'' is often displayed when it
stops, the most relevant aspect of auto-regression is the type of ``externalized'' feedback that
it provides for the system. And this is an essential part of its functioning. Note that in no
step of the architecture so far did ChatGPT store any information, there is no ``internal''
state, no memory. It is, in a sense, a purely reactive system\footnote{You can explore this
yourself by asking GPT to imagine a number or retain something secretly and perform operations on
it and the like.}. It is through auto-regression that it does compensate for it, in a manner
that will become very important when addressing the agentive capacities of GPT.

The term transformer was originally chosen to depict the sequential transformation of the input
matrix into an output without recurrence, with the task of translations as a key component
\citep{vaswani2017}. It was later applied to other tasks (like summarization) and finally
discovered that a large enough transformer could perform very well generally across tasks. And
also that it could ``learn'' what to do simply by direct instruction with none or few examples
of what was asked to do \citep{brown2020,radford2019}. This is when the concept of prompt takes
significance. The power of LLM transformers is so general and unspecific that it is open to be
prompted to unfold in different directions: summarization, translation, correction, explanation,
conversation, expansion of key ideas, development of outline strategies, etc. The ``magic'' so to
speak that sustains these capacities, lies on the parameters of the system, the embedding,
attention and FFNN matrices (coloured purple on Figure~\ref{fig:gpt3}) that operate on the input
matrix.

We know very little of GPT-4 and subsequent OpenAI models. Despite its name and the original
goals of the company soon after GPT-3 it moved into secrecy, releasing very little information
of new models, other than some raw details and benchmarking results \citep{openai2023}. It is
speculated that one of the greatest architectural innovations of GPT-4 over its predecessors was
the inclusion of Mixture of Experts. This is a technique of splitting the network into
specialized subnetworks and training it to learn to distribute processing to them
\citep{shazeer2017}. We directly know about the effectiveness of this technique, because it has
been effectively implemented on open source or and more transparent LLMs that match GPT4
benchmarking performance like Mixtral \citep{jiangAQ2023}, LLaMa \citep{metaai2025}, KIMI K2
\citep{moonshot2025} or DeepSeek \citep{deepseek2025}. The other important innovation is
multimodality. Tokenization has nowadays turned multimodal. Images are broken into a grid of
patches (e.g., 16$\times$16 pixels) and each patch treated like a ``word'' (a token), thus
creating a sequence of ``image tokens'' that can be fed into the main Transformer
\citep{dosovitskiy2021}. The same goes for audio waveforms. Once encoded, all these different
tokens (text, image, audio) co-exist in the shared, high-dimensional vector space, where
relationships and patterns can be learned between modalities\footnote{Multimodal generation is
more complicated. Instead of recursively adding tokens to a string, like language models do,
image, sound, or video synthesis typically relies on generative \emph{diffusion models}
\citep{ho2020}. They work on what can intuitively be understood as a kind of ``internalised
sketch pad''. The model starts with a tensor of pure random noise and iteratively refines it by
subtracting the noise until a clean image emerges. In multimodal text-to-image systems, this
``denoising'' process is conditioned on the text prompt, steering the generation towards an
output that is ``semantically'' aligned with the user's description.}. Although functionally
relevant, these innovations do not imply a radical change from the explanation we just provided.
The architecture remains fundamentally the same.

%% ----------------------------------------------------------------
\subsection{Training}

GPT and other LLM configuration is typically carried out in various stages. The first is,
somewhat paradoxically, called ``pre-training'' but constitutes the main training (understood as
the process by which one improves or acquires new capacities). During this process, the
parameters of each processing transformation just described gradually change until a given level
of accuracy is reached, pre-training ends, and they remain fixed until new training procedures
start. Then comes fine-tuning, with two basic stages: task specific fitting and reinforcement
learning. Finally, prompt learning is often used, which is more of an instructional form of
directing the system.

\subsubsection{Pre-training}

Explaining first the way of functioning of the whole architecture, as we just did, is essential
to understand training. Contrary to other approaches, each processing block is not trained in
isolation to perform a specific task (e.g.\ 1\textsuperscript{st} grammatically articulate the
input, then build a general abstract representation, next, carry out inferences and take an
output decision), but the entire system is trained at once, through \emph{back-propagation}
\citep{rumelhart1986}.

The basic mechanism is simple: the system is initialized with random parameters. Next, a chunk
of input (e.g.\ the beginning of a sentence) is chosen among a training dataset. It is then
processed as we explained above. This is called a forward pass. This pass finishes when the
system provides the result array: that which indicates the probabilities of all the words to be
the next one (the step before selecting the final output). The result will be nonsense at the
beginning. For example, to the input ``Elephants don't play'' the highest probability of the
result array could be ``purple'', followed by ``Fodor'', ``misuse'', ``chain'', ``cat'', etc.
Now, this is compared to the correct result: an array that gives 0 probability to all the words
except 1 for ``chess''. But ``chess'' might be very down on the assigned probabilities. Yet, it
is now possible to compute an error (or loss): the difference between the assigned probabilities
on the result array and the target one.

Next, this loss will be backpropagated through the network (the backward pass). By means of an
optimization algorithm, small changes are made all throughout the whole network in the direction
of minimizing this error: The algorithm calculates the response to the question ``what change
should I do to this parameter so that the resulting output reduces the error?'' and makes the
change accordingly, for each parameter on each block, backwards.

This process is iterated once and again, until the forward process produces no or little errors.
All three major components of the LLMs are trained in this way: embeddings, attention mechanisms
and feed-forward networks. Although the overall procedure is locally relatively simple, the
amount of little changes is vast and the effect is the performance capacity we can witness today.
The computational cost of training GPT-3 was 3.14$\times$10\textsuperscript{23} FLOPs, that is,
314 sextillion floating-point operations \citep[Appendix~D]{brown2020}.

The system so far is considered a raw \emph{foundation model}: it can process text generally and
can be put to work on a number of tasks already, or be further trained to improve performance on
specific types of tasks. The training process, so far, is considered unsupervised, nothing other
than the next-word-prediction is used to train the model.

\subsubsection{Fine-tuning and Reinforcement learning}

Additional training procedures are used to fine-tune the transformer for specific tasks, like
summarization, translation, or conversation. This time the instruction (e.g.\ summarize) and the
task input (e.g.\ a whole Wikipedia article) are provided, and the system is trained with
back-propagation to match a model output (e.g.\ Wikipedia's summary entry for that article),
instead of just the next token. This is considered \emph{supervised learning}, no human
intervenes yet, but the task is not simply to ``guess'' the following word but to match a
specific target goal, pairs of input and target-output are required to complete this training.

Transformers are usually further trained to include \emph{Reinforcement Learning with Human
Feedback} or RFHF \citep{ziegler2020}. The pre-trained and fine-tuned LLM is let to interact
with humans. Then, based on how humans have positively or negatively evaluated the output of the
model, it is trained to produce outputs that are more likely to be positively rated or less
likely to be negatively valued; according to the past corrections made by human interactors. This
is where the system is often trained on ethical or moral values, together with a number of other
quality checks.

Finally, we have \emph{one-shot or few-shot learning} procedures that operate basically at the
prompt level, providing examples or specific instructions that the LLMs take as input to produce
new examples or follow the instructions provided \citep{brown2020}.

\subsubsection{Self-Improvement and AI-Supervised Training}

In order to overcome the limitations of LLM to solve complex problems (e.g.\ in maths or
coding), new innovative training techniques have been introduced. Interestingly, they often
involve the use of AI generated feedback, data, or supervision. In Process-supervised Reward
Models (PRM), for example a specific artificial model, the PRM, is first trained with a huge
database of reasoning paths where every step has been labelled by humans as correct or incorrect.
The first goal here is to train the PRM to evaluate good reasoning, assigning rewards to each
individual reasoning step. The LLM is then trained by the PRM, acting as a ``teacher'' via
reinforcement learning (RL) to produce novel reasoning chains given new prompts
\citep{lightman2023}. This would be equivalent to teaching first a machine to identify good steps
on cooking recipes, and then using it to teach a cook to generate good recipes by judging each
step to achieve the final (prompted) request.

Other methods involve, instead of reinforcing specific steps on a reasoning chain, to
differentially reward different chains of thought according to their relative quality
\citep{shao2024}. In this case, the student is left to generate different recipes and is rewarded
according to how good each recipe would comparatively result to a culinary critic (who doesn't
really know \emph{how} to cook, but has been trained to develop a good taste). This is the
formula behind the success of DeepSeek-R1 \citep{deepseek2025}. A further innovation from this
company involved distillation, where a large, powerful ``teacher'' model is used to generate the
vast synthetic dataset of high-quality reasoning paths and ranked-choice answers\footnote{The
DeepSeek team generated one order of magnitude more reasoning chains artificially for distillation
(about 800K) than the original dataset presumably used by the OpenAI team with 75K examples
\citep{lightman2023}.}. The smaller, more efficient ``student'' model is then tuned on this
machine-generated corpus (similar to the pre-training phase). Following the metaphor, once a good
cook is trained, it is possible to ask it to generate millions of new recipes and to train a
smaller (cost-efficient) model with them, overcoming the limited amount of recipes in available
cookbooks.

These AI-supervised procedures can be applied not only to improve reasoning, but also to
``induce'' ethical guidelines. This is the case of \emph{Constitutional AI}, where the model
learns to align its behaviour with a set of written principles, a ``constitution'', defined by
humans, but does so without direct human feedback. The process involves an initial phase where
the model is prompted to critique and revise its own responses according to the constitutional
principles. The system is then tuned to predict the next token of those self-revised responses.
In a second phase, a teacher preference model is trained on the constitutional principles and then
RL with AI feedback is applied to the LLM \citep{bai2022}.

%% ----------------------------------------------------------------
\subsection{Anthropomorphising GPT}

Calls to avoid anthropomorphizing GPT are recurrent
\citep{bender2021,butkus2020,coeckelbergh2021,jebari2021,kubes2022,shardlow2023}. But
anthropomorphizing is here only referred to as projecting human qualities, particularly cognitive
or emotional ones, to the machine. Something that is perceived as a risky strategy, since
understanding (or experiencing) the interaction with ChatGPT through the human or intentional
stance \citep{dennett1989} as if it truly had genuine human capacities, would make us falsely
attribute a set of properties it certainly lacks. Properties that are essential to the human
social world-making: commitment, trust, responsibility, empathy, etc. Important as it is, the
emphasis of this type of anthropomorphisation shadows other important forms. There are at least
two more types of anthropomorphisation that are relevant to understand GPT\@. And their analysis is
perhaps more revealing of its mode of existence than the attribution of mental properties to the
system. The first such type of anthropomorphizing is the way in which the training corpus and
procedures shape the machine as a human. The second is the inverse process of trying to bring to
the human scale and capacities the internal workings of the system. We shall attend to both in
this section in a somewhat combined manner.

Regarding the processing, according to \citet{kaplan2020}, we can roughly approximate the
computational cost of processing a single token to be directly proportional to the number of
parameters of the model. GPT-3 having 175 Billion parameters, the computing cost of writing a
250 token summary of a 1750 token essay could have the approximate cost of
2000$\times$175 Billion = 350 trillion FLOPs (floating point operations)\footnote{Although the
recursive nature of the output processing could indicate an exponential growth of this cost, it
is effectively reduced by not re-processing the whole input again (see this discussion for a more
detailed explanation \citealt{tunstall2022}).}. Carried out by a human, as an experientially
graspable task, each FLOP could be approximated as equivalent to a multiplication between two
5-digit numbers. Assuming such an operation could take about 10 seconds to be completed by an
expert or well-trained human being\footnote{It actually took one of us a few minutes to complete
it!}, it would take around 500 million years of human labour, working 40 hours a week, to process
that prompt\footnote{At the same time, it is worth noting that the human brain, at a subconscious,
subpersonal level of activity, can carry out the equivalent of this 350 trillion FLOPs in about
one second or less \citep{carlsmith2020}.}. That means that John Searle would have to live and
die \emph{a few million times} before he could output even the first symbol from his Chinese room.
Intuitions that once worked at a certain scale (like the Chinese room experiment) might not
necessarily be trusted at many orders of magnitude higher scales.

Regarding the training aspect, we know little of GPT-4 but GPT-3 was trained with 570 GB of
text data, about 300 billion tokens according to their own creators \citep{brown2020}, that is
approximately 200 Billion words. Thus, the training data for ChatGPT is equivalent to about 2
million books. A volume so vast that it would take a human being more than 500 years to read
through it all, assuming they dedicated 8 hours a day, every day, reading at a speed of 200
words per minute. To put this into further perspective, if an average person reads roughly 500
books in their lifetime, the amount of data ChatGPT has been trained on is comparable to the
combined reading of 4000 lifetimes. But if we were to humanly compute all the backpropagation
process of 314 sextillion FLOPS, that would take an expert human 4.19$\times$10\textsuperscript{17}
years to compute, which is almost 7 orders of magnitude (30,386,783 times) the age of the
universe.

Let's now focus on the inverse analysis of how ChatGPT is already anthropomorphized by the
training data, including the biases, themes, styles and poetic tendencies that are present in
them. And not less importantly, by all the fine-tuning and human reinforcement learning. To say
it differently, ChatGPT has no way of organizing tokens around mothers, mice, or forests other
than that provided by human traces on texts. It is, thus, not surprising that we can
anthropomorphize it. It already is. And it is so in a manner that cannot be fully grasped. Unlike
a mannequin that we can touch and verify that its shape is indeed human, but whose functioning is
nothing more than that of a piece of inert plastic. We can not even bring the complexity of the
concrete functioning of GPT down to a graspable human scale. Its intensive and enormous training
procedure, its gigantic internal structure and its vast mode of operation lies beyond the human
scale of understanding\footnote{So does a bacterial cell or the global economy, by the way; not
to mention the human mind and the extended socio-cultural scaffolding.}. However, its internal
and behavioural functioning can be generally (if not specifically) sufficiently understood so as
to determine constraints to its mode of existence. And we can properly ground why and how we can
avoid ontological anthropomorphization. Agency being a pivotal, often anthropomorphized, category
for settling the ontological status of LLMs.

%% ----------------------------------------------------------------
\subsection{Towards LLM based agents}

At a first sight, nothing in this architecture qualifies properly as agency. Not even for the
most optimistic or naive engineers. The system is fully driven by the prompt and directly steered
when output completion has taken place by a new prompt. Moreover, the system has no internal
states, no (internal) memory, no potential desires, goals, or purposes. When operating (after
training is completed), not even a trace of what is processed is left within the system; except
for the history of outputs that is continuously fed-back into the system auto-regressively. In a
sense, GPT operates like Leonard Shelby, the protagonist of Christopher Nolan's celebrated film
\emph{Memento} \citep{nolan2001}. Devoid of the capacity to create new memories (yet able to use
its knowledge), Leonard externalizes instructions (goals, instrumental steps, etc.) and
contextual information (pictures, notes, etc.) to regain the agency that he lost due to his
amnesia.

The lack of agentive capacities of the raw GPT is apparent in the type of digital embodiments
that AI engineers are providing to enhance GPT and develop so-called ``autonomous GPT agents''
\citep{andreas2022,huang2024,wang2023,weng2023,xi2023,zhouW2023}. March to June 2023 saw a
rapid increase of projects trying to deploy digital agents based on GPT and other LLMs: AutoGPT
\citep{significant2023}, AutoGen \citep{wuQ2023}, DemoGPT \citep{unsal2023}, SuperAGI
\citep{admin_sagi2023}, MiniAGI \citep{mueller2023}. A number of initiatives have followed that
promise to deliver fully operational agents for programming \citep{wuS2024,yang2024} and tech
giants seem to be betting on LLM-driven agents to make generative-AI services profitable
\citep{holmes2024,knight2014}.

There are 5 kinds of LLM enhancement strategies that are being developed to move from ChatBots
to the so-called ``agents'': a) extended memory systems, b) planning strategies, c) reflexive
evaluations, d) the use of tools, and, e) multi-agent interactions. These strategies are most
often implemented in combination, but it is, nevertheless, possible to differentiate them.

\emph{Extended memory} frameworks (like Langchain) make it possible for transformers to
temporarily extend and sediment autoregressive dynamics (e.g.\ rewriting and organizing summaries
of past input context to increase its memory and better focus it on specific task-goals).
Sometimes such extensions are not different from our practice of externalizing memory in a
notebook, writing To-Do lists and offloading planning structuring in a bullet point document.

\emph{Planning strategies} involve prompting the transformer to split a specific goal or task
into sub-operations that can then perform in sequence. This can be achieved through various
techniques, the most known of which is the so-called Chain of Thought or CoT \citep{wei2023}.
CoT is implemented by crafting prompts that encourage the model to ``think aloud''. Instead of
trying to answer directly to a given question, or to accomplish a task, the LLM is first prompted
to explicitly write down how it will plan to do it and then follow its own plan. This technique
has been shown to enhance LLM reasoning and planning abilities. More sophisticated methods, like
Tree of Thought \citep{yao2023,zhouA2023} involve combining a tree-like decomposition of a
variety of plans with LLMs capacity to reflexively evaluate the adequacy of each potential plan
(which brings us to the next point).

\emph{Reflexive evaluation} procedures as simple as asking the transformer to reflect on the
previous output and correct existing mistakes have been shown to dramatically increase
performance \citep{madaan2023b}, also to provide some degree of self-guidance on the completion
of the decomposed sub-tasks. The generative and creative capacity of LLMs to deliver execution
plans is often combined by using LLMs to automatically evaluate them and to distill a more
consistent strategy. Both planning and reflexive evaluation are often now built-into ``reasoning''
LLMs (as explained in sec.~3.2.3).

\emph{Use of tools}: LLM can be connected to a wide variety of tools
\citep{mialon2023,openai2025b,schick2023,shen2025}, from programming consoles like Python, to
web-browsers, search engines, and, more generally APIs (Application Programming Interfaces) that
make possible to interact with digital services through instructions (rather than visuomotor
interfaces). These ``tools'' define the ``bodies'', interfaces and environments of LLM powered
``agents''. In November 2024, Anthropic launched the Model Context Protocol to define such tools
and interfaces for LLMs, and it has rapidly expanded as a standard to let LLMs access different
services \citep{anthropic2024,hou2025}.

\emph{Multi-agent interactions}: Finally, in order to overcome memory limitations, lack of
consistency or repeated failure, multi-agent approaches are used, which involves interacting,
evaluating and selecting results from other agents \citep{liJ2024}. Increasing successful task
completion through collective agency is frequently achieved by combining many of the techniques
explained above, like self-organizing Tree of Agents strategies \citep{chen2024}.

Despite the increasing enthusiasm on the potential of LLMs to provide solid foundations for
digital agents, strong limitations have already surfaced: LLMs do not seem to be much better
discriminating than they are generating plans \citep{jiangD2024}, they are very limited on their
capacity to develop complex plans \citep{kambhampati2023}, perhaps because it is still very hard
\citep{wuW2024} for LLMs to integrate future tokens on their current processing; which is a
fundamental way in which humans plan.

In Chat scenarios, human intervention can continuously stir the conversation, discard
hallucinations, or ignore wrong answers. Agentic scenarios are different. The human presence in
the conversational domain makes the coupled LLM-human system much more fault-tolerant. But when
humans are out of the loop, errors can accumulate catastrophically. Think of the cumulative
effects of hallucinations or mistakes on making a cake: a mixture of eggs shells, salt, and flour
could end up in the fridge instead of the oven, despite a 98\% accuracy on the design and
execution of the recipe. It is thus no surprise that, unlike benchmarks directed at measuring
linguistic capacities, intelligence or knowledge, LLMs still score far behind humans in current
agentic benchmarks \citep{liu2023,valmeekam2023b,xieJ2025,xieT2024}. It is therefore still early
to judge whether they can at least operate ``as if'' they were genuine agents. This remains an
open empirical issue. Meanwhile, it is possible and necessary to explore how existing transformer
architectures meet the requirements for agency identified at a more fundamental level than that of
pure performance.

%% ================================================================
\section{LLMs are not (autonomous) agents}

We have seen how LLMs based on transformer architecture internally operate and how their
capacities have been expanded with a series of additions to the foundational trained models.
Moving below the surface of performance-level measurement to characterize agency requires a
certain commitment to theoretical or philosophical frameworks. It is essential to specify the
nature of actions, purpose, and cognitive properties, and to relate them to the underlying
generative mechanisms. The detailed technical account of LLMs' architecture and agent-like
extensions will turn out critical to assess their ontological status. In this section, we first
approach the assessment from the point of view of computational representationalism (from which
LLMs can comfortably be characterized as agents). We then move to alternative, so called, 4E
frameworks, whose requirements severely problematize agency attribution to LLMs.

From the philosophical perspectives that have given credit and have contributed to the AI
research program, it is difficult to rule out genuinely agentive capacities from ChatGPT-like
systems. Representational computationalism is one such approach
\citep{carruthers2006,newell1980,putnam1965}. It is a type of functionalism that defines mental
properties (intelligence, knowledge, learning, or agency) in terms of the input-output functional
(internal transition) states of a system representing states of affairs of the environment. The
essential feature of the mind is the capacity to reason or to draw inferences upon representations
of the world; i.e., information processing. For instance, you take the umbrella because you just
read it will be raining today, and you know that umbrellas are a good way to cover yourself from
the rain. According to representational functionalism, this is the kind of inference that is
characteristic of mental processes. And, LLMs are well capable to make such inferences. Moreover,
their internal states reasonably approximate the world. Being models of language, and being
trained on a huge amount of text, to the extent that all these training data can be squeezed to
provide a model of the world, LLMs, are also models of the world
\citep{kadavath2022,liH2023,yildirim2024} including other agents \citep{andreas2022}\footnote{According
to this view, having no ``real'' contact with the world is no fundamental obstacle. Certainly
(some) LLMs have no vision capabilities to see if it is raining right now, but nor do you, when
you read in the newspaper or your favorite weather-app that it is about to rain. The interface of
information reception does not affect the nature of the inference that it is appropriate to bring
the umbrella with you. In turn, a LLM, without a robotic body, could not itself complete the
action to take the umbrella, but it could perfectly command you to do so (by means of a text
message) or could signal the cars' top window's controller to close it. The nature of the
cognitive process of making the right inference according to the right knowledge of the world is
indifferent to the mediation of the input or output. In this sense, LLMs could be considered
full-blown cognitive agents with more or less sensory and motor capabilities.}.

Some authors have gone even further, proposing that all reality is informational and agency is
the ability to act upon and be affected by the (informational) environment. Agency is thus not
limited to human beings but can also apply to artificial entities such as robots, software
programs, and AI systems.

\begin{quote}
``These new agents already share the same ontology with their environment and can operate in it
with much more freedom and control. We (shall) delegate or outsource to artificial agents
memories, decisions, routine tasks and other activities in ways that will be increasingly
integrated with us and with our understanding of what it means to be an agent.''
\citep[p.~62]{floridi2007}
\end{quote}

From this perspective, LLMs could be considered agents perfectly embedded in the infosphere. In
fact, Floridi has recently contemplated this possibility and the problems and challenges involved,
concluding that GPT-like AI systems are ``agency without intelligence''
\citep{floridi2023,floridi2020}. His category of agency can be understood as depending on two key
components: capability and autonomy. Capability refers to an agent's ability to perform a certain
action or set of actions, while autonomy (for Floridi, and much of AI engineering) refers to an
agent's ability to act independently, without being controlled or directed externally. In short,
this theoretical framework can be summarized as follows: (1) all entities are informational, (2)
some (informational) entities are agents, and (3) agents are entities that perform actions
independently of one another.

However, this characterization of agency might be too liberal. Agents are characterized by their
capacity to carry out actions, as distinct from mere events or mechanically caused states of
affairs\footnote{Even when it is considered that actions are caused by events \citep{davidson1980},
these are of a very special kind: reasons, beliefs, desires, etc.}. Actions, unlike (other)
events, are not merely occurrences in the world (informational or otherwise); they are processes
imbued with intentionality and purpose. This distinction becomes evident when comparing the
experience of intentionally reaching for a bike, an action, with being inadvertently pushed
towards it by the wind, an event. The former is characterized by a sense of directiveness and
intention, elements central to the phenomenology of agency we experience and recognize every day.
Any output of a system (computing machine or otherwise) does not automatically qualify as an
action.

Many have questioned the adequacy of informationalist and computationalist approaches to capture
and explain cognitive and agentive capacities. In this critique converge theoretical contributions
from different fields: phenomenology \citep{gallagher2017,merleau1944}, philosophical and
theoretical biology \citep{jonas1966,maturana1980,moreno2015}, philosophy of mind
\citep{searle1980,noe2004,hutto2012,thompson2010} empirical contributions from the psychology of
perception \citep{gibson1979,heras2019,reedES1996} or conceptual development \citep{lakoff1980},
large-scale neuroscience \citep{buzsaki2006}, methodological contributions from complex dynamical
systems' theory \citep{barandiaran2006,favela2020,port1995}. These and other criticisms have
resulted in a family of alternative approaches that are often labeled under the term 4E-cognition
\citep{gallagher2023}; standing for embodied, extended, enactive and ecological.

Within these, the approach outlined by \citet{barandiaran2009} allows for the comparison of
natural agency with the operations of LLMs (or any other system)\footnote{For a similar approach
that identifies requirements and limitations of AI systems to meet enactive standards for
cognition see \citet{froese2009}.}. They start by reviewing different available definitions of
agency (from software engineering to robotics, from philosophy to psychology) to bring together a
surface description of what these definitions have in common: ``a system doing something by
itself according to some goals or norms''. They spell out what this commonality entails,
identifying 3 necessary and sufficient conditions for agency: individuality, normativity and
interactional asymmetry. First, an autonomous agency requires that a system be self-individuated.
Second, the self-individuation process defines a set of norms (of viability) and, third,
according to these norms, the system asymmetrically regulates its coupling with the environment
(thus becoming the source of an action). In sum, from their perspective, an agent system is an
autonomous organization capable of adaptively regulating its coupling with the environment in
order to sustain itself according to the rules set by its own conditions of viability
\citep[p.~376]{barandiaran2009}.

A bacterium moving up a sugar gradient \citep{berg2004} is a widely accepted paradigmatic
example of agency that satisfies the definition \citep{barandiaran2014}. First, the bacterium is
in a continuous process of individuation and self-distinction. Metabolism produces the components
of the reaction networks constituting the agent, together with a membrane that separates the
system from its environment. This self-production in turn determines which aspects of the
environment are relevant, normatively valued (good or bad) from the very constitution of the
system: some chemical compounds are essential nutrients for its self-maintenance (good), some
others are poisonous compounds that degrade the membrane or the metabolic network (bad). Finally,
the agent modulates its coupling with the environment by moving up or down a sucrose (positively
valued nutrient) gradient and absorbing sugar molecules across the membrane. The whole combination
of self-individuation, norm generation, and adaptive regulation constitutes the agentive nature of
the bacterium's behavior.

Processes of individuation and normative regulation need not happen at the metabolic scale
exclusively. Mental or sensorimotor life can also ground agentive capacities
\citep{barandiaran2007,barandiaran2008,dipaolo2017}, bringing it closer to our own experience of
intentional agency. Not only do we experience our living body (our biological agency), but also
our actions in the world are guided by goals and intentions that transcend mere biological values.
This is so because a new level of autonomous organization emerges through the neural mediation of
behavior. The individuation process is constituted by a self-sustaining network of sensorimotor
schemes (e.g.\ habits) in continuous development\footnote{For more representation minded
approaches, this can also be conceived as a network of beliefs whose main behavioral manifestation
is the growth and maintenance of the network. Free energy and active inference approaches are
relatively consistent with this view and several parallels have been drawn with enactivism
\citep{clark2013,kirchhoff2018,seth2021}, although severe objections to identify both theories
have also been drawn \citep{aguilera2022,dipaolo2022,nave2025,raja2021}.}.

Our experience of agency stems from the fact that we are a mesh of habits that, through specific
actions, asserts its own identity. We shape ourselves as behaving systems by acting. A
psychological and cognitive identity develops through activity dependent plasticity, organizing
brain, body, and environment. And the norms that emerge from this identity direct my behavior. I
identify myself as a philosopher, I want to make a good contribution analyzing GPT, I struggle to
write these lines correctly. The goal of a specific task is nested into a network of interests and
plans that ultimately rest on preserving our identity \citep{barandiaran2025b}.

Yet I can be coupled with my environment in many ways that, despite involving myself and my norms
or goals, do not qualify as action. Someone else, chance, or simple physical constraints, might
move or prevent certain moves so as to contribute to my norms, e.g., a nurse in a hospital taking
care of me. But the result of this behavior would not qualify as agency yet (in fact, I am a
patient in a hospital, not an agent). It is not until the source of my behavior lies asymmetrically
laden to my psychological or sensorimotor capacities and my sensitivity to my self-generated
norms, that my coupling to my environment can properly be called an \emph{action}.

Does ChatGPT meet the three criteria for agency? Let's analyze them one by one. The first
condition, \emph{individuality}, requires that a system self-produces or at least self-sustains,
distinguishing itself from an environment it co-defines. LLM's existence and maintenance, however,
are reliant on external human intervention and tools, and it operates within a predetermined
environment. This diverges from the self-individuation process essential for autonomous agency.
Note that it is difficult to tell from the agentive perspective what is the system in GPT\@. On
the one hand, there is the input, transformed into a matrix and processed through the complex
network of operations described in section~3. On the other is the transformer that is nothing but
a set of operator blocks, blindly transforming the input and the resulting matrices. Underneath is
the hardware, whose operations are indifferent to the type of processing taking place, or even to
the fact that no processing takes place.

When considering \emph{normativity}, an agent system is inherently at risk of degradation without
specific actions, and sensitivity to these viability conditions that emerge from its process of
individuation is essential for normative behavior. ChatGPT, however, operates without such
precariousness and does not autonomously establish its goals. Its functions are programmed
externally, driven by an error or loss function that is completely extrinsic to the operations of
the system. This absence of intrinsic normativity is evident in ChatGPT's inability to recognize
failure autonomously, often leading to repetitive or non-productive responses. A counterintuitive
property of LLM is that they are typically capable, retrospectively, if explicitly asked, to
identify ``errors'' on their previous operations \citep{madaan2023b}. And yet the absence of
implicit normativity is apparent in that, given their autoregressive character, they never
``realize'' their mistake unless prompted to do so.

Instead of purposeful action (behavior that is imbued with purpose all along its unfolding)
transformers are somehow partially purpose-\emph{structured} as a result of the pre-training on
the massive purpose-expressive texts, and purpose-\emph{bound}, that is, statistically falling
within ``humanly interpretable normative'' bounds, as a result of fine-tuning and supervised
learning procedures. Moreover, training procedures used to shape autoregressive loops as ``think
aloud'' strategies for so-called ``reasoning models'' endow them with purpose-structured and
purpose-bounded generative procedural capacities\footnote{In a sense, it is possible to identify
these systems as second and third order derivative intentional systems, following the classical
distinction between \emph{intrinsic} or genuine intentionality, characteristic of humans, and
\emph{derived} intentionality, characteristic of signs and artifacts \citep{haugeland1990,searle1983}.
LLMs get their derived intentionality from that of their training data (second-order derived
intentional systems) and, in turn, generate intentionally-structured products that recursively
shape (through chain-of-thought techniques) their subsequent outputs (third-order derived
intentionality). A more detailed account of this connection between Generative AI and
intentionality can be found in \citealt{barandiaran2025a}.}. However, unlike living agents,
ChatGPT lacks the essential concern or awareness for goal achievement that converts an operation
into a purposeful act.

The concept of \emph{interactional asymmetry} underscores that an agent system is the originator
of its actions, modulating its interaction with the environment becoming the \emph{source} of the
action. ChatGPT, however, exhibits a fundamentally reactive nature (it has no internal state). It
responds passively to external inputs (prompts) and, as we just identified, lacks self-defined
norms to guide its interaction with the environment. Pushed by an initial prompt, its operation,
however complex, rolls down a fixed (albeit probabilistic) and instantaneously reactive path
toward the next output. And yet, the autoregressive aspect provides a powerful form of recursion
that, within the right context, particularly that provided by ``agency'' extensions of LLMs,
increases the interactional asymmetry property towards the transformer. When carefully crafted,
LLMs can partially escape their downhill reactive nature (by rewriting the landscape they roll
through) but, devoid of intrinsic individuation and normativity, fail to become genuine agents.

When measured against the criteria for autonomous agency derived from contemporary embodied and
enactive frameworks, LLMs fall decisively short. This conclusion is grounded in the specific
technical and operational details we outlined in Section~3. Their lack of self-sustaining
individuality, their operation based on externally defined loss functions and reinforcements
instead of intrinsic normativity, and their fundamentally reactive nature are core architectural
and operational limitations that preclude genuine agency.

Once established what LLMs are not, we are now faced with the constructive task of defining what
they are. The next section, therefore, moves beyond critique to propose a positive conceptual
framework for understanding their unique mode of existence and their transformative potential for
human agency.

%% ================================================================
\section{What are transformers then, if not agents, and how do they transform agency?}

%% ----------------------------------------------------------------
\subsection{The language automaton and the ghost in the human-machine interaction}

If transformer architectures, and LLMs as we know then, are not (autonomous) agents\ldots\ What
are LLMs that have such a strong impact on our digital environment and the way we live in them?
If we are not to embrace transformers as members of our familiar way of being (as agents) in the
world, we need to start somewhere else. First, it might be useful to distinguish operations from
actions. An \emph{operation} is a sequence of occurrences that can be interpreted functionally;
that is, in (instrumental means-ends) relation to a final end state. Mechanisms carry out
operations. A \emph{digital operation} is a logical transformation executed by a computer.
Operations are externally defined, whereas actions are internally defined by the agent that
carries them out (in the sense outlined in the previous section). Note that an action can
externally be defined as an operation and translated into a machine. But it is more convenient to
use a specific name to label those processes that can be described both as an operation when
carried out by a machine and as an action when done by a human: we might call those \emph{tasks}
(see Table~\ref{tab:conceptual}).

\begin{table}[htbp]
\centering
\caption{Conceptual divide between machine and human types of identities, properties, and types
of behavior. The middle column captures a common vocabulary that permits both humans and machines
to share an interactive conceptual space. Note that systems can combine machine-individual hybrids
to become individuals and individuals are (also) systems and can perfectly be interpretable as
performing automatically.}
\label{tab:conceptual}
\smallskip
\begin{tabular}{>{\centering\arraybackslash}p{0.26\textwidth}
                >{\centering\arraybackslash}p{0.26\textwidth}
                >{\centering\arraybackslash}p{0.26\textwidth}}
\toprule
\textbf{LLM} & \textbf{either} & \textbf{Human} \\
\midrule
Machine      & System              & Individual  \\
Operation    & Performation/Task   & Action      \\
Automata     & Performer           & Agent       \\
Automatic    & Performant          & Autonomous  \\
End state    & Goal                & Purpose     \\
             & Calculator          &             \\
             & Writer              &             \\
             & Painter             &             \\
             & Interlocutor        &             \\
\bottomrule
\end{tabular}
\end{table}

With these distinctions at hand, we can now proceed to properly characterize LLMs. From the point
of view of their organization, transformers are \emph{automata}, as opposed to autonomous
systems. This distinction is still relevant and crucial. Automatic\footnote{Although the etymology
of automatic or automata ultimately brings us to self-minded or self-willed with a very strong
mentalistic connotations, its popularization to depict mechanisms capable of completing
sophisticated chains of operations is the sense that it nowadays embodies.} systems do not need
human intervention to carry out certain operations, but are not autonomous. They cannot define
their own identity and norms. However, as structured and identifiable digital instruction sets in
physical memory and processors, they can carry out complex sequences of operations in the real
world. They can transform energy into operations without human supervision or intervention during
the process. They are \emph{automata}.

However, transformers are not any kind of automata, they are of a very special digital kind, and
operate in a very special type of environment, with an even more singular relationship to human
life: they are \emph{digital language automata} operating in multiple language-supporting and
language-driven digital networks we continuously inhabit as linguistic animals (together with many
other digital and physical objects around us).

Moreover, LLM-powered chatbots, like ChatGPT, are specifically constituted by the way they couple
with their associated milieu: other humans. As such, they often become \emph{phantasmic language
automata}\footnote{Our use of metaphors such as `ghost' and `phantom' is a deliberate
methodological choice. In periods of rapid technological change, where the conceptual landscape
is unsettled, metaphors serve as essential cognitive and philosophical tools for structuring an
inquiry into a novel phenomenon \citep{hesse1966,lakoff1980}. They function by inviting us to
adopt a new perspective or ``framing'', which organizes our understanding of a subject in ways
that a literal description cannot \citep{camp2006}. This allows us to take seriously the
phenomenological character of the human-LLM interaction, which a purely technical description
might obscure. Metaphorical phrasing allows us to condense intuitive understandings, anticipate
future lines of inquiry, and reorganize our conceptual framework in ways that can precede and
guide more detailed technical articulation. Our aim is to use these metaphors as conceptual
probes, which we then ground in the technical and philosophical analysis that follows. In fact,
the public and academic discourse surrounding LLMs is already deeply shaped by metaphors, from
the deflationary ``stochastic parrots'' to ``autocomplete on steroids''. Our goal is to provide
new metaphorical framings that we believe more adequately capture the complexity and
transformative potential of these technologies. In this case, the metaphor of a ``phantasmatic
language automaton'' captures a dual nature. It is an automaton because its operations are
mechanically determined and devoid of the self-generated norms characteristic of genuine agency.
The `phantasmic' aspect, however, arises not from the machine alone, but from the human user's
phenomenological co-creation of the interaction. Humans readily project and enact a conversational
partner when presented with a sufficiently coherent interlocutor. The LLM serves as a perfect
scaffold for this projection, creating a `phantom' conversational-other that is experientially
real for the user but lacks an autonomous source on the machine's side.}. In some sense, ChatGPT
is certainly the \emph{ghost} of the text corpus that it originally was trained on. As a
``phantom'', GPT can talk to us without the capacity to bring itself to life. Nevertheless, it is
capable of conjuring all the knowledge of the corpus, of which it is a shadow. As an automaton,
it is capable of producing changes as a result of a sophisticated mechanism. Changes that trigger
the response of the acknowledgment of an equal from those with which it phantasmatically
interacts. In this sense, it is possible to characterize ChatGPT as an \emph{interlocutor
automaton} in a double sense that mirrors the twofold meaning of the term ``interlocutor'': a) as
a system capable of performatively interacting in effective conversations with humans (and with
other machines traditionally designed to take human produced text as input, e.g., programming,
database queries, etc.); and b) as an intermediary between (practically all digitized textual)
human knowledge and other humans or machines.

In this sense, ChatGPT operates as a gigantic \emph{text-that-talks}, or rather a
\emph{library-that-talks}\footnote{And, we could say, in the more advanced multimodal cases, as a
media-library-that-talks-and-paints.}, enabling a dialogical engagement with the vast corpus of
human knowledge and cultural heritage it has ``internalized'' (compressed on its transformer
multidimensional spaces) and that it is capable of recruiting effectively in linguistic exchange.
The machine's interlocution, though devoid of personal intentionality, bears the trace of human
experience as transposed into digitized textuality. The purpose-structured and bounded automatic
interlocution, however, can be experienced as a genuine dialogue by the human subject. As a
result, ChatGPT brings all this digitized textual knowledge, within the technological milieu,
\emph{ready-to-chat}.

Readiness to chat also implies \emph{readiness-to-command}. And since we live in a world of
linguistic performance and greatly digitized linguistic milieu, commanding can easily be turned
into performance by a linguistic automaton. This is where recent enthusiasm with LLM ``autonomous
agents'' lies. Transformers (properly ``embodied'') can command themselves and other LLMs. Unlike
mechanical automata, linguistic automata can be commanded by goals expressed in natural language,
and can, at least partially, evaluate whether the results of the tasks carried out match the
linguistic goal-expression. The capacity that truly brings linguistic automata close to some
aspects of human agency is that commands of this sort need not be expressed within the strict
boundaries of a computationally interpreted language (like programming languages or shell
commands). LLMs can operate within the flexible and context dependent mesh of ``natural''
language.

Yet, the phantasmic dimension of LLM chatbots does not only reveal itself in its capacity to
bring the dead text into non-living automatisms. It also has to do with ChatGPT as an enacted
other, when given the interlocuting role. To appreciate the complexity of our interaction with
ChatGPT, it is first essential that we understand what happens when we speak with other
autonomous linguistic agents and what happens when we speak with ``phantoms''. Embodied and
enactive approaches to sociality \citep{dipaolo2018,gallagher2017,perez2021} defend that social
cognition involves dynamic interaction with others. Social cognition is not about rationally
reconstructing the thoughts that others are holding, but the result of ongoing fluent interactions
between two or more agents. It has fundamentally more to do with dancing with another person than
rationally strategizing a chess play by mail. Conversations are constituted by a complex chain of
partial acts that imply, and somewhat anticipate, their completion by others (e.g., giving and
taking, question and answer, salutation and response, etc.). What happens when the other is
absent? Well, in a sense, we often play as-if it was there. We imagine that it is there, and
re-enact a completion of our acts (e.g., an imaginary conversation with a friend). In a sense,
we incorporate the absent other, internalizing the various dimensions of what would otherwise be
an open interaction.

The experience of a phantom is somewhat a continuation of this capacity, to which we add the
perceptual (visual, sound, etc.) hallucination as a means of a partial externalization of the
experience. With ChatGPT, we have \emph{excorporated} the phantom. The phantom is ``real'', but
still a phantom. The perceptual feedback is real (the text and sounds we hear come really from
out there) but some hallucinatory and self-completive aspects of the interaction are still
constitutive of it. There is no real-agent on the other side, but we still act as-if there was
one, thus in a sense making it a real conversational experience \citep[see also][]{persson2024}.
Similar phenomena have been reported in Human-Robot Interaction research, describing the human
tendency to navigate a `social artifact puzzle' by treating mechanical systems as intentional
agents to make interaction more tractable \citep{ziemke2023}.

%% ----------------------------------------------------------------
\subsection{Transformer embodiments}

Cognitive science has turned from abstract symbolic computations into the (historically neglected)
role of the material body in the production of mindful experience and capacities
\citep{calvo2008,gallagher2023,shapiro2019,varela1991}. Cognitive agency is said to be embodied,
extended beyond the brain as the ``mere'' hardware of the mind executing the genuine mindful
``inmaterial'' software, into the living body of the cognizer, its technical equipment and
sociomaterial environment. There are many layers of embodiment that Transformers rest on, and
many that they lack compared to those of human intelligence. Most notably, ChatGPT lacks a
sensorimotor body that captures variations on its physical environment (it lacks physical sensors)
and also motor actuators that change its relationship with the environment and induce further
sensory changes \citep{chemero2023}\footnote{Modern multimodal systems identify and correlate
patterns across modalities but lack the active, world-directed and normatively endowed nature of
embodied agency. The crucial feedback loop, where an agent's actions generate new sensory
contingencies that shape subsequent actions, remains absent. The model analyses a static image,
but it does not visually and actively explore a normatively valued environment to achieve an
intrinsic goal. Thus, multimodality expands the model's input beyond pure text, but it does not
provide a sensorimotor body in the enactive sense. It merely expands the ``library'' of passively
acquired data into a ``media-library''.}. But, as we mentioned in the introduction, a
sensorimotor robotic body is quickly being integrated with existing multi-modal LLMs, and its
impact on deeper linguistic and behavioral skills can be expected to be significant. However, so
far, transformers lack a living and lived body (real or simulated), that is often associated with
emotions and affectivity regarding cognitive or agentive capacities
\citep{damasio1994,thompson2010}. This certainly sets humans apart from LLMs. But there is
nothing new or specific to LLMs on such lack of embodiment. In the previous section, we have
sketched some consequences of this not-being-alive. This is a notable difference between
Artificial and Natural intelligence that has attracted attention and arguments elsewhere as well
\citep{froese2009,koch2019,seth2021,thompson2010,ziemke2009}. We shall now focus instead on two
forms or aspects of LLM embodiment that are novel and have received little attention from the
point of view of agency.

As a first type of embodiment (although in a sense very different from the usual one), we should
focus on the \emph{written corpus} (body in Latin) on which LLMs are trained and that they so
effectively bring back to text. This is not simply a metaphorical use of the term body or
embodiment. Textuality is an abstract, yet very concrete and complex, form of materiality itself,
an externalized embodied product of linguistic agency. It brings with it purpose- and
experience-structured relationships that might bear deep isomorphism with them. As human
digitized culture is partly accessible to us, we often neglect the tremendous value (and size)
of LLMs compressed \emph{corpus}: an organized model of the textualized human knowledge and
culture\footnote{We are so immersed in the complexities and subtleties of our culture and so
``shocked'' by the power of the new AI that we frequently only focus on what still distinguishes
us from it or assign purely instrumental value to it. However, how would we qualify the value of
a LLM-like device, were it the only or primary source of access to an extinct human culture (or
a distant alien one) for which we (or the LLM) possess (nevertheless) some kind of translation
capacity?}. We need a theory of what this written corpus really is from an embodied and
phenomenological cognitive science perspective. But current theories, so far, lack an account of
the cognitive or agentive implications of large scale computable textuality; and of the
transformation of social, cognitive and ultimately biological lifeforms they bring with it.

For some theorists, it could have perfectly been the case that a new generation of AI-systems was
bootstrapping itself to human level intelligence by means of pure rational deduction and
interaction with the environment (perhaps also as the exclusive result of the engineering effort
of a private corporation\footnote{Ideally, also, that corporation produced or acquired the
information and environment for the robot to become smart and also assumed the cost of
externalities associated with its AI's intellectual growth.}). That is not the AI we have
available today. This one is built on the collaborative effort of thousands of mathematical and
computer science contributions, it is fed or trained on huge amounts of universal written
heritage, collaborative digital commons (like Wikipedia or massive open source code repositories)
and millions of distributed conversations and cooperative efforts (mostly on internet forums). No
less important are the \emph{embodiments of care} that usually take the form of labor
externalization of massive data curation and operational alignment supervision, and that requires
and recruits social-emotional resources from underpaid labor to safe-rail the brute models
\citep{perrigo2023,xiangC2023}. Another significant aspect of LLMs embodiment is their heavy
computational and thus energetic and resource-hungry nature, and the extreme capitalist
supply-chain extractivism it triggers, demands, and sustains \citep{valdivia2024}.

In this sense, more than a self-bootstrapped Artificial Intelligence, ChatGPT, as an interlocutor
automaton, is a computational proxy of the human collective intelligence externalized into a
digitalized written body. It is, in turn, shaped and taken care of by hundreds of human and
non-human lives. Thus, although in a very wide sense, yet one that is crucial to the effective
operation of LLMs, transformers are embedded on large scale human and ecological bodies. This
happens not just at a contextual level or as an operational environment, but at a constitutive
level. No LLM is an island. And their performative power, and derived agentive capacities (if
any), inherently rest on human and planetary scale life.

These dimensions of the material embodiment of LLM training and execution are crucial to
understand the types of asymmetries it will bring to extended agentive capacities in humans.
During the last 40 years, with the advent of the PC, there was little extended-agency computational
asymmetries between human agents in general. Access to specific types of information has always
been asymmetric between humans, but, beyond this, the informatic capacity of a 14-year-old hacker
and that of a big corporation was relatively even. With some rare exceptions, you and I could do
the same thing with a PC or a mobile device compared to what Elon Musk, Queen Elizabeth or Warren
Buffett could do. The computing (energy, processing, and data) cost of compiling and executing the
most complex of software products (e.g., an operating system) was relatively accessible to most.
Now, although (relatively low cost) access to the best generic LLMs is still ``widely''
guaranteed, and despite locally-executable LLMs' quality is growing, the asymmetry on the capacity
to train and deploy specific LLMs, is orders of magnitude bigger.

On the one hand, there is the power implicit on how and what to train LLMs on, a power in the
hands of those very few with the resources to train and shape a foundation model: the direct
constraining power of training data choices and training procedure selection, the power of
establishing RLHF criteria, constitutional writing capacity and interface design of massively
deployed LLMs. On the other hand, the so-called ``AI autonomous agents'' only operate with some
efficiency under exponential computational costs, either by a ``social'' distribution of tasks,
by massive parallel planning and evaluation or by deeply iterative and long ``chains of thought''.
That is, by making up for the lack of genuine intrinsic purposefulness with redundancy. Their
effective deployment might bring super-agency to privileged human masters, while delivering
low-quality, low-cost alternatives to most. The asymmetry that the subsequent power amplifies is
enormous. This is certainly not unlike the recursive power asymmetries that are already present in
contemporary societies\footnote{The richer you are, the higher your capacity to hire good lawyers
to reduce your tax payments, your capacity to pay consultants to make more profit from your
investment, etc.} except for a crucial fact: language digital automata remove humans (and their
capacity for disobedience, resistance, and uprising) from the (social) sources of power.

%% ----------------------------------------------------------------
\subsection{Assisted, extended and midtended agency}

What is ChatGPT as an extension of human cognitive agency? Is it an \emph{assistant}? Most
technologies that have been thought of as extensions or embodiments of human activity have been
thought of as bodily prostheses. The extended mind hypothesis \citep{clark1998} and its later
developments, including material engagement theories \citep{malafouris2016}, have focused on the
way in which beyond-the-skull extended material or computational processes should be understood as
constitutive of cognition or brain processes. According to this view, mindful thinking processes
must often be understood as extended into the material environment that they shape and, in turn,
also shapes cognitive processes. Thinking involves \emph{thinging}. When we make pottery, we
don't print or carve an internal mental 3D model into the clay, we mold it. The materiality of
the clay guides us, we bring the jar into form through a continuous reciprocal interaction between
brain, body and world. Digital technologies constitute a branch of these phenomena: we offload
memories, drafting procedures, image manipulation, etc.\ into our PC and mobile devices. Cognitive
gadgets are organs of our mind extended beyond the skull.

Social interaction also extends and assists human agency: crew and team members interact
(coordinatively or subordinatively), to achieve levels of agency (collective or directive) that
are unreachable to a single individual \citep{lewis2022}. The interlocutionary capacity of LLMs
brings human autonomous agency to a level not-unlike that of intersubjectively augmented agency,
where the machine takes the role of an assistant (or that of a master). Recent human-computer
interfaces have been dominated by action directed design. We tell computers what they have to do
(by programming a specific function, by pushing a button, by dragging a file, by selecting a menu
item). LLMs, instead, make it possible to prompt or command an intention \citep{nielsen2023}: ``I
want a summary of this text in French so that the 5-year-old child of my visiting friend can
understand it'', ``I need an impressionist style picture for a book cover on philosophy of mind''.
This is certainly going to bring us to unprecedented forms of \emph{master-slave} dialectical
relationships with machines (and the corporations they ultimately serve). On the one hand, we
might soon have at our disposal an ``unlimited'' number of assistant automata ready to perform
computational and linguistic tasks for us. On the other hand, we might be equally commanded by
them\footnote{A step further on the already widespread tendency to be systematically commanded by
apps running on real time data and AI driven optimization of work (like Uber, Glovo, etc.).}, not
only through textual interfaces but also more sensorimotorly embodied throughout the course of our
everyday behavior (e.g., Meta glasses connected to LLMs capable of interpreting our visual scenes
and delivering ``suggestion'' or ``orders'' to accomplish specific tasks \citep{meta2023,waisberg2023}).

But Transformers are also bringing with them a much deeper meaning of extended agency (with deeper
dialectical connotations). There is a form of extended agency that LLMs already offer that becomes
more intentionally intimate than any previous known form. In fact, this \emph{ex}tensional
character is closer to the \emph{in}tentional character of the mind that deserves a proper name:
\emph{midtent}ional\footnote{The `mid-' prefix is chosen to situate this process in the middle
ground between purely \emph{internal} sourcing of intentionality, and classical \emph{extended}
intentionality, where cognition is offloaded onto passive environmental resources.
\emph{Midtention} describes the hybrid process that emerges when an external system, itself
possessing generative capacities (like an LLM, or another human), actively and reciprocally
intervenes in the agent's ongoing stream of active externalized thinking (like writing, designing,
composing, coding, etc.) resulting on products that the autonomous agent takes as her own; yet
the intentional process has been somehow displaced to an inter-mediate(d) space between the agent
and a generative assistant. For a detailed account of this concept of ``midtended cognition'' see
\citealt{barandiaran2025a}.}. We might best illustrate this form of mixed enhanced agency with
some type of LLM integration on programming and office environments. But let's first stop to
analyze the phenomenology of non-AI-assisted writing. The process of writing (in paper or on the
screen) is one that is often experienced as the very act of writing driving itself the intentions
of the writer: the interaction process of writing pulls agency out of the head. It is the recurrent
hand-keyboard-PC-screen-vision-brain-body-hand loop that produces text. Yet, we don't only write.
We also supervise and edit recurrently. Thus, at least two loops are involved here, one is more
pulled by the direct writer-text dynamics, the other by the more detached editorial supervision
that either continuously or intermittently follows the former. At times, one finds the non
consciously written text as right and owned, as proper, and it is left untouched. Other times\ldots\
``That is not what I meant exactly'' ensues, ``it needs a rewrite''. Both loops are person
anchored. The environment (pen and paper or keyboard and screen) served as a support structure, a
well integrated, creative scaffold, providing the material basis of extended memory, recomposition,
and tinkering. But the writer was the extended agent, the organized center of the scaffolded
subject. This might start to change.

The enormous complexity and regulatory capacity of the brain-body system (compared to that of the
passive materiality of the tool and work environment) is now challenged by an ongoing activity of
language automata, which are constantly reading us and writing (for) us. The extended autocomplete
experience that LLMs provide can be tuned to integrate previous documents and styles of the
writer, it mobilizes background knowledge, and is context-sensitive and purpose-structured (almost
as if your shadow could push you into the direction you are intending to move towards). By feeding
the LLM with the input tokens of the collaboration between its past output tokens and those written
by you, the autocomplete feature adapts as you type. It often provides a mixed sense of agency,
where what you find pre-written in the screen is ``yours'' without it having been necessarily
written by you\footnote{In computer coding, this brings the expression of ``predictive coding'' to
a completely new level.}.

This brings the power of transformer-human interaction closer to a proper \emph{cyborg agency},
beyond any experience of instrumental, social or intersubjective agency we might have ever
encountered before (for a detailed account of cyborg intentionality, albeit pre-GPT, see
\citealt{verbeek2008}). In a textualized manner, this form of autocomplete is equivalent to
injecting predictive efferent signals into the body movement. It is a step far beyond the
classical examples of extended mind, in which, despite an out-of-the-skull spread of cognitive
processes (the notebook, the mobile memory storage, the pen-and-paper calculation), the complexity
asymmetry and integrative capacity was always tilted towards the skull-side of the coupled
agent-environment system. It implies a degree of intimate technical transparency
\citep{andrada2023,perezverdugo2022} in generative activities that challenges the very nature of
human agency.

If we take predictive processing theories at face value \citep{clark2013,clark2023,friston2009,seth2014,seth2021}
we might be encountering, for the first time, that the environment is delivering to the brain-body
the very predictions that the brain-body is about to make about the effect of its own activity on
the environment. It is tempting to consider this as a short-circuiting of agency as we know it.
And it is difficult to assess the full consequences of the improvement of this technology, its
massive adoption and multimodal expansion (introducing, perhaps, a new chapter on the many ethical
implications of cognitive technologies, see \citealt{clowes2024}). This brings the interlocutionary
nature of LLMs to an \emph{intra}locutionary mode of existence that uncannily blends with us.

%% ================================================================
\section{Discussion and conclusions}

The irruption of LLMs on our digital world, and through it on our lives (digital or otherwise),
is breaking (again) what we thought human artifacts could never do. They perform operations that
would require high levels of complex, common-sense, unstructured, and creative intelligence if
performed by humans. We are forced to question their ontological status and the deep way in which
they can transform ours. In this paper, we have focused on responding to this question, focusing
on agency.

We started identifying the polarized take on LLMs ontological status: from their inflationary
characterization as fully ``sentient beings'' to the deflationary one as mere ``stochastic
parrots''. Next, we delved into the faulty yet outstanding capacities that LLMs display, as
measured by different human-level intelligence benchmarks. We concluded that a deeper delving
into their organization is necessary to properly determine their mode of being. A detailed
explanation of the complex and powerful architecture and processing of LLMs was provided next,
together with the training and tuning procedures used to shape them. We also explained the
different techniques that are used to enhance these systems to achieve agentive capacities.
Turning to contemporary embodied philosophy of mind made it possible to identify what is missing
from current models for them to achieve the status of genuine autonomous agents. We next moved to
make a positive proposal as to how to treat LLMs and the way they transform our agency. It is
time now to wrap up some concrete conclusions out of this journey.

%% ----------------------------------------------------------------
\subsection{Autonomous agents, interlocutionary automata, and deeply embodied midtention}

You are enmeshed in a thick web of recurrent attention-intention loops, of which you are both
cause and effect. Through the growth and arrangement of this web, you have developed a sensitivity
to navigate and stir your behavioral world so as to care for its deep precariousness. Along this
open process, what is continuously guiding your action is not the anticipation of the next token
of a pre-given text (or data-stream), but the sensitivity to the consequences of your actions at
different nested scales: on the task at hand, on your goals, plans, and, ultimately, on your own
identity (itself the result of your own actively sustained and stirred encounters within the
world). You are a genuine autonomous and organic agent. LLMs are not. But they perform,
historically-unprecedented, extraordinary tasks. And they will continuously intersect with the
way in which we navigate our (linguistic) worlds.

If by autonomous agent we mean an automatic system that is efficient on a sequence of multiple
tasks, then LLMs (with important extensions and within structured digital environments) might
already deserve the name. If, by agency, we mean the sense of agency we experience as autonomous
self-defining and self-governing systems, then there are good reasons to believe that the LLM
architecture as we know it currently falls short to meet the demands. This might not be a bug but
an intended safety feature of transformer architectures. Systems that display complex intentional
capacities might be a powerful assistant at a high price. Autonomous agents of this kind are the
most powerful and yet most dangerous assistants. Being capable of creating your own norms, and
being adaptively capable of displaying complex strategies to meet and transform them, is
compatible with accepting external commands and making external norms your own. But it is also
open to revolt. And this, in turn, opens a whole set of problems of AI alignment and safety.

A fundamental question remains: is it possible to achieve efficient and automatic multi step
task-completion without genuine agency? Some strong requirements (like deep material living
embodiment) might never be met by transformer-like systems. But it is still possible to envision
variations on the current architecture that could bring the system's operations closer to living
actions.

The deep transformation we are witnessing bears some relevant parallels with the industrial
revolution and the increasing factory automation. Different degrees of automation work well within
factories, thanks to the operational constraints provided by the assembly line (always under human
supervision and care). This time, the internet is the assembly line (rather network) for
linguistic production. A LLM, put into the right sociotechnical environment (its ``associated
milieu'', \citealt{simondon2017}), becomes an \emph{interlocutor}, a task \emph{performer}, or
even an \emph{operator} of the language fabric and the visuomotor internet (accessible through a
web browser and keyboard-mouse actuators). The machine's interlocutionary capacity is real when
coupled to other systems. It is certainly devoid of personal intentionality, but it brings humans
to conversational life (as a clinical language automaton capable of supporting life without being
itself alive). It is capable of this by means of the complex (context-sensitive, non-linear and
massively parallel) re-generation of the multiple traces of human experience that the corpus of
digitalized textuality embodies. Devoid of purposefulness, the intensive data training and
human-machine guided reinforcement procedures make LLMs, however, purpose-structured and
purpose-bounded systems that can command and be commanded.

When intimately coupled with human digital activity (and its history), LLMs can augment existing
forms of agency in various ways. Some are well captured by the concept of ``assistant'' and
involve forms of agency enhancement similar to those achieved by social coordination or
subordination. Other forms of extending human agency are new. They involve the intervention of
LLMs on the very ongoing activity of the human agent, by anticipating (based on context and
previous history) the next token action(s) recursively. We called this \emph{midtended} agency,
in which machine operations blend with the subject into a unifying intentional process.

In order to understand the new technological landscape that LLMs open, we don't need (yet) to
sacrifice the distinctive character of our autonomous agentive capacities. But we need to gain a
detailed understanding of the capacities and dialectical processes that such systems will trigger.
Language automata are here to stay, and we need to tune our conceptual systems to accommodate
them.

%% ----------------------------------------------------------------
\subsection{On the dangers of the ``stochastic parrot'' metaphor and the ``agency without
intelligence'' conceptualization}

Deflationary accounts of AI tend to forget that human agency can, at a very fundamental level of
quantum mechanics or the less fine-grained level of neuronal modelling, be characterized simply
as a collection of dumb ``probabilistic'' and ``stochastic'' processes. It is not the description
of the basic mechanisms that compose a complex system what defines its properties, but the
organization of interrelated processes (both internal and interactive) that such basic mechanisms
make possible to emerge. This is as true of us (living humans) as it is of any machine. In order
to assess the genuine capacities of a system, we need to look at their internal workings, and the
emergent capacities they can display when organized and coupled in specific manners. We need
careful conceptual crafting to approach systems in which the relationship between the description
of local mechanisms and the display of interactive capacities spans so many orders of magnitude
(around 20) that we can hardly grasp.

LLMs are nothing alike stochastic parrots, nor domesticated living animals nor stochastic engines,
neither caged or free in the rainforest, but effectively coupled to the digital fabric of our
social life. Whereas the metaphor of the ``stochastic parrot'' was once useful to question the
rapidly emerging hype on LLMs \citep{bender2021}, it might easily turn counterproductive. Parrots
are living agents, ecologically balanced within their habitats, and capable of actively adapting
to environmental changes (including those induced by human capture). They achieve this by means of
deep cognitive, emotional and communicative capacities (far beyond the traditionally attributed
dumb mimicry; see \citealt{pepperberg2006}) that LLMs certainly lack. On the other hand, LLMs
display capacities that effectively mobilize human intelligence as embodied in massive textuality,
affectively mobilize human intelligence in conversation, and can activate forms of hybrid agency
previously unavailable for human intelligence. And they do so by displaying powers far beyond
those of stochastic, probabilistic or merely statistical token recombination.

Functionalist or informationalist conceptualizations don't play better than the ``stochastic
parrots'' metaphor in the sociotechnical jungle. They fail to distinguish autonomous agency from
mere digital processing. Declaring LLMs as ``agents without intelligence'' does not fix the
foundational failure, it simply highlights it by unveiling the impotency to properly justify lack
of intelligence on transformers. It also reverses the ontological order. They are better
understood as (collective) intelligence without agency. The inversion does not only ignore the
increasing material and energetic demands of AI and fuels LLM corporate marketing discourses. It
misplaces our own agency and responsibility.

%% ----------------------------------------------------------------
\subsection{Prospects for transformed agencies in the era of deep digitality}

Some of the properties that are essential to agency (most notably individuality and normativity)
emerge from the deep materiality of organic agency. However, the recent course of AI explosion,
with the gigantic investment of data and computational capacity (and the related energy demands)
is revealing a \emph{deep digitality} whose consequences are still to be fully unpacked. The
complexity and scale of the operations involved in LLMs training and execution are huge. Prompt
processing operations that, carried by an aware and conscious human, would take billions of years,
challenge our intuitions and conceptual resources. By a digitality that deep, it is reasonable to
hold that the boundary between invention and discovery, between artifact and nature, between
engineering and science is somewhat blurred. We (humans) have built LLM as much as we have
discovered their emergent capabilities\footnote{In fact, it is important to note how LLMs are
rarely said to be built but ``trained''.}. And avenues for a genuine digital agency might still be
open for discovery. The way in which deep and wide materiality has revealed agentive capacities in
natural history might well be somewhat replicated in the digital realm. If deep materiality brings
with it the capacity to make difference emerge \citep{anderson1972} we have no reason to preclude
the increasingly deep digitality of artificial devices to reveal new forms of agency, yet to come.
And even deeper transformations of the existing forms of agency.

But depth alone does not bring matter (or digitality) to life. It is ultimately the organization
of processes, their interaction with their environments, their interdependence with the rest of
beings, that needs to be scrutinized to disclose the mode of existence of any device. No benchmark
or general description (stochastic, statistical, probabilistic, syntactic, or otherwise), is
sufficiently informative of the potential transformative capacities of machines. LLMs are no
exception. Their mode of existence is highly dependent on human (and other) forms of life. And
the deeper our materiality and digitality merge, the deeper will be the transformations to come.
This is why transparency and openness regarding LLMs (and AI in general) is much more than a
private ethical imperative and turns into a collective political concern: how these systems work
and get coupled to our social fabric, on how they feed on the human heritage and care, how they
suck planetary resources and affect social inequalities. To shape this future, we need a better
conceptual understanding of how the mode of existence of LLMs transforms real agency.

%% ================================================================
%% Bibliography
\bibliographystyle{apalike}
\bibliography{transforming_agency}

\end{document}